\documentclass[11pt,letterpaper]{article}
\usepackage[margin=1in]{geometry}
\usepackage{lmodern}
\usepackage[round,authoryear]{natbib}
\usepackage[utf8]{inputenc}
\usepackage[T1]{fontenc}
\usepackage[hidelinks]{hyperref}
\usepackage{url}
\usepackage{booktabs}
\usepackage{amsfonts}
\usepackage{nicefrac}
\usepackage{microtype}
\usepackage{xcolor}
\usepackage{amsmath,amssymb,amsthm,mathtools}
\usepackage{graphicx}
\usepackage{subcaption}
\usepackage{enumitem}
\usepackage{array}
\usepackage{tabularx}
\usepackage{float}
\newcommand{\E}{\mathbb{E}}
\newcommand{\PP}{\mathbb{P}}
\newcommand{\Corr}{\mathrm{Corr}}
\newcommand{\Cov}{\mathrm{Cov}}
\newcommand{\Var}{\mathrm{Var}}
\newcommand{\Bin}{\mathrm{Bin}}
\newcommand{\1}{\mathbf{1}}
\newcommand{\cM}{\mathcal{M}}
\newcommand{\cP}{\mathcal{P}}

\newcommand{\dd}{\,\mathrm{d}}
\newcommand{\Vinf}{V_\infty}
\newcommand{\ME}{\mathrm{ME}}
\newcommand{\Bern}{\mathrm{Bernoulli}}

\theoremstyle{plain}
\newtheorem{theorem}{Theorem}[section]
\newtheorem{proposition}[theorem]{Proposition}

\theoremstyle{remark}

\theoremstyle{definition}

\title{Two Calls, Two Moments, and the Vote-Accuracy Curve of Repeated LLM Inference}
\author{Yi Liu\\[3pt] {\normalsize York University}}
\date{}

\begin{document}
\raggedbottom
\maketitle

\begin{abstract}
\textcolor{black}{Repeated sampling can improve LLM accuracy, and quantifying the gains from additional calls is essential for allocating test-time compute. We study binary decisions, a fundamental setting where repeated answers to the same question are aggregated by majority vote. We show that two independently sampled responses per example in a validation set with known answers constrain the latent distribution of example-level success probabilities to a class consistent with the paired outcomes. Optimizing over this class yields sharp accuracy and gain bounds at every finite voting budget. A shared large-sample confidence region accounts for validation uncertainty. We also obtain sharp infinite-vote bounds and moment-matched forecasts of the vote-accuracy curve. On QNLI and QQP, our method successfully distinguishes settings with voting gains above a target margin from those with little room for improvement.}
\end{abstract}
\section{Introduction}

\textcolor{black}{Repeated sampling has become a standard way to spend test-time compute. Self-consistency samples multiple reasoning paths before aggregating final answers \citep{Wang2023}; pass@$k$ evaluation measures success across repeated code-generation attempts \citep{Chen2021}; and inference-scaling work studies repeated candidate generation as a performance axis \citep{Brown2024}. A parallel line uses consistency and entropy of repeated generations as uncertainty evidence \citep{Manakul2023,Kuhn2023,Farquhar2024}. These methods raise a statistical question: for a fixed model, prompt, and sampler, how much extra accuracy can majority voting extract from more calls?}

\textcolor{black}{The difficulty is that consistency is not correctness. Repeated calls may agree because the system reliably solves an example, or because it reliably makes the same mistake. Conversely, a model with lower single-call accuracy may retain randomness that voting can average away. LLM outputs can vary even under nominally deterministic serving settings \citep{Atil2025}. With black-box access, evaluating voting on a new task requires repeated responses and additional inference compute.}

\textcolor{black}{We study binary decisions, the simplest setting for this question, with applications including LLM safety review \citep{Han2024WildGuard}. Each call answers the same question, and voting succeeds when a majority of calls are correct. Our analysis applies to any stochastic binary predictor whose calls on each example are independent with a common success probability, which can vary across examples. Averaging majority-vote success over examples gives the vote-accuracy curve: accuracy as the number of calls increases. This curve says whether three, five, or more calls improve on one.}

\textcolor{black}{Estimating the full distribution of success probabilities would require many repeated responses per example. Two calls already constrain the vote-accuracy curve: on a validation set with known answers, we record the correctness of two responses per example. The single-call success rate identifies the mean success probability; the rate of two correct responses identifies its second moment. These rates also determine the correlation of correctness across calls on the same example. The paired outcomes turn a potentially many-call validation problem into a two-moment problem: distributions with the same moments can yield different vote-accuracy curves. Their smallest and largest accuracies at each budget give the sharp bounds.}

\textcolor{black}{The central reduction makes this problem tractable. For every finite voting budget, optimizing over all success-probability distributions with these two moments reduces to distributions supported on three points and matching quadratic upper and lower bounds. Each endpoint is attained by a compatible distribution, making the bounds sharp. The same reduction bounds the gain from increasing an existing voting budget. A shared large-sample confidence region for the moments supports simultaneous budget comparisons while accounting for validation uncertainty. We derive closed-form three-vote bounds and sharp infinite-vote bounds. Two moments also support useful point completions when a single curve is desired. Maximum-entropy and Gaussian-probit models fitted to these moments provide full-curve forecasts.}

\textcolor{black}{Experiments on QNLI and QQP show why single-call accuracy does not order the benefits of repeated inference. Highly correlated calls can leave little room for improvement, while other settings yield positive lower bounds on voting gains. In held-out evaluations, the two-call analysis establishes gains exceeding half a percentage point. An observed ranking reversal further shows that a model with lower single-call accuracy can overtake another after voting. These results guide decisions about starting voting and increasing the number of calls.}

\section{Related work}
\label{sec:related-main}

Repeated-sampling methods differ in their aggregation targets. Self-consistency aggregates sampled reasoning paths \citep{Wang2023}; adaptive-consistency uses agreement-based stopping \citep{Aggarwal2023}, while confidence-informed self-consistency weights answers by model confidence \citep{Taubenfeld2025}. Code-generation and inference-scaling studies evaluate repeated attempts through pass@$k$ and larger sampling budgets \citep{Chen2021,Brown2024}. \textcolor{black}{Pass@$k$ measures the chance of at least one successful attempt. Plurality selects the most frequent answer, so its success also depends on how incorrect answers split.}

Agreement-based uncertainty methods use repetitions as evidence about reliability. SelfCheckGPT compares sampled responses for hallucination detection \citep{Manakul2023}; semantic uncertainty and entropy group linguistically different but equivalent answers \citep{Kuhn2023,Farquhar2024}. Empirical studies examine how uncertainty proxies discriminate errors and calibrate confidence \citep{Savage2025,Vashurin2025,Xiao2025}, while ConU calibrates prediction sets for correctness coverage \citep{Wang2024}. Repeated-prompt studies emphasize the dependence of measurements on the same example \citep{Gallo2025}, and \citet{Nowak2025} analyzes the allocation between distinct prompts and repeats when estimating self-consistency.

The C-bound controls weighted-majority risk through the first two margin moments, with \textcolor{black}{PAC-Bayesian estimation guarantees} \citep{Germain2015}. \citet{Liu2026Signature} characterizes the complete odd-budget voting curve through a signed signature. \textcolor{black}{Moment-constrained optimization uses small-support extreme points of moment sets} \citep{Winkler1988,Pinelis2016} and polynomial majorant/minorant duals \citep{KarlinStudden1966,Bertsimas2005}. Maximum entropy \citep{Jaynes1957} and normal-ogive item-response models \citep{Albert1992,Lord1980} \textcolor{black}{select latent laws.}

\section{Repeated calls and two-moment information}
\label{sec:model-main}

To formalize the validation probe, fix a model, inference protocol, target population, and correctness evaluator $e:\mathcal Y\times\mathcal Z\to\{0,1\}$. For ground truth $Y_i$ and possibly structured call output $Z_{ij}$, the evaluator includes parsing, normalization, and an invalid-output rule. \textcolor{black}{The score $B_{ij}$ equals one for a correct response and zero otherwise.} Let $q_i$ be the success probability on example $i$, let $\pi$ be its law across target examples, and write $Q\sim\pi$ for a generic draw:
\begin{equation}
B_{ij}=e(Y_i,Z_{ij}),\qquad
q_i\stackrel{\mathrm{i.i.d.}}{\sim}\pi,\qquad
B_{ij}\mid q_i \stackrel{\mathrm{ind}}{\sim}\Bern(q_i).
\label{eq:hierarchy-main}
\end{equation}

Repeated calls are independent conditional on $q_i$, although marginally dependent because they share the example. Independent sampling units are example draws, not individual calls. A fixed distribution over prompts is also a fixed inference protocol when every call samples its prompt afresh and normalizes the answer to the same task label. Model or checkpoint changes define different protocols. Hidden serving-state reuse, feedback between calls, or correlated decoding streams need not satisfy \eqref{eq:hierarchy-main} (see Appendix~\ref{app:dependence-stress-main}).

\textcolor{black}{For $K=2n+1$ calls, let $P_n(q)$ be the probability that a majority is correct on an example with success probability $q$. Averaging over examples gives the population accuracy $V_n(\pi)$:}
\begin{equation*}
P_n(q)=\sum_{\ell=n+1}^{2n+1}\binom{2n+1}{\ell}q^\ell(1-q)^{2n+1-\ell},
\qquad
V_n(\pi)=\int P_n(q)\,\dd\pi(q).
\end{equation*}
Thus $V_0=\mu=\E Q$ is single-call accuracy and $V_1$ is \emph{three}-vote accuracy. Fair binary tie-breaking makes an even budget equal in accuracy to the preceding odd budget (Appendix~\ref{app:even-collapse-main}). Selecting the first answer on a binary tie is also conditionally fair; counting ties as incorrect changes the objective.

The binary scores need not come from a binary output vocabulary. When outputs are normalized to one of two task labels, ordinary label majority has accuracy $V_n$ and needs no ground truth at deployment. For structured outputs, $V_n$ instead denotes the event that a strict majority passes the evaluator. With a unique correct answer this event guarantees plurality success, so its lower bound transfers to plurality, but its upper bound does not. The two-sided decisions below apply only to a deployed rule whose accuracy is $V_n$.

\begin{proposition}
\label{prop:moments-main}
Under \eqref{eq:hierarchy-main},
\[
\E[B_{i1}]=\E[B_{i2}]=\mu:=\E[Q],
\qquad
\E[B_{i1}B_{i2}]=\nu:=\E[Q^2].
\]
For $0<\mu<1$,
\begin{equation}
\rho=\Corr(B_{i1},B_{i2})=\frac{\nu-\mu^2}{\mu(1-\mu)}\in[0,1].
\label{eq:rho-main}
\end{equation}
The feasible set is exactly
\begin{equation*}
0\le \mu\le1,\qquad \mu^2\le\nu\le\mu.
\end{equation*}
\end{proposition}

Appendix~\ref{app:proof-moments-main} proves the moment identities and constructs a latent law for every feasible pair.

The pair distribution gives the operational meaning of the moments:
\begin{equation}
\PP(B_{i1}=1,B_{i2}=1)=\nu,
\quad
\PP(B_{i1}\ne B_{i2})=2(\mu-\nu),
\quad
\PP(B_{i1}=0,B_{i2}=0)=1-2\mu+\nu.
\label{eq:pair-probs-main}
\end{equation}
\textcolor{black}{The disagreement probability} $2(\mu-\nu)=2\E[Q(1-Q)]$ measures within-example randomness; the covariance $\nu-\mu^2=\Var(Q)$ measures between-example heterogeneity. Together they decompose the variability of a single correctness score:
\[
\mu(1-\mu)=\Var(Q)+\E[Q(1-Q)].
\]
At a fixed mean, large $\rho$ means that more variability is tied to the example and less remains across calls on that example. Successes and failures are then more reproducible. \textcolor{black}{Section~\ref{sec:exact-main} quantifies the disagreement needed to guarantee voting gains.}

\textcolor{black}{Writing $\cP([0,1])$ for all probability distributions on $[0,1]$, the latent laws compatible with the two-call distribution form the moment class}
\begin{equation}
\cM(\mu,\nu)
=
\left\{\pi\in\cP([0,1]):
\int q\,\dd\pi(q)=\mu,\quad
\int q^2\,\dd\pi(q)=\nu
\right\}.
\label{eq:momentclass-main}
\end{equation}
The population bounds below hold uniformly over $\pi\in\cM(\mu,\nu)$. For a budget of $2n+1$ calls, the sharp accuracy interval is
\begin{equation}
L_n(\mu,\nu)=\inf_{\pi\in\cM(\mu,\nu)}\int P_n(q)\,\dd\pi(q),
\qquad
U_n(\mu,\nu)=\sup_{\pi\in\cM(\mu,\nu)}\int P_n(q)\,\dd\pi(q).
\label{eq:LnUn-main}
\end{equation}
Sharpness means that no smaller interval covers all compatible latent laws. \textcolor{black}{With exact moments, compatible laws can still give different accuracies; finite validation adds sampling uncertainty.}

\section{Sharp vote-accuracy bounds}
\label{sec:exact-main}

With only a mean $\mu$, the law $(1-\mu)\delta_0+\mu\delta_1$ \textcolor{black}{(where $\delta_x$ is a point mass at $x$)} remains compatible with the data and gives $V_n=\mu$ at every budget. Thus one call cannot guarantee a strictly positive gain. Appendix~\ref{app:mean-only-main} gives the exact mean-only three-vote envelope.

\textcolor{black}{To see what the second moment adds, compare this population with one in which every example has success probability $\mu$. Both have single-call accuracy $\mu$. In the first, every example is either always correct or always wrong, so voting leaves accuracy unchanged. In the second, $Q=\mu$ and three-vote gain is $P_1(\mu)-\mu=\mu(1-\mu)(2\mu-1)$, which is positive for $1/2<\mu<1$. Their second moments are $\nu=\mu$ and $\nu=\mu^2$, respectively. Two calls distinguish these explanations of the same average accuracy; the sharp bounds extend this comparison to every intermediate feasible moment pair.}

\subsection{Three-support reduction and quadratic dual}

The class \eqref{eq:momentclass-main} includes arbitrary discrete, continuous, and mixed latent laws, so optimizing over it is initially an infinite-dimensional problem. The next theorem shows that each best- or worst-case vote accuracy is attained by a population with at most three latent success probabilities. Quadratic envelopes establish the corresponding endpoints over the entire class.

\begin{theorem}
\label{thm:general-odd-main}
For every feasible $(\mu,\nu)$ and every fixed $n\ge1$, \textcolor{black}{both extrema in \eqref{eq:LnUn-main} are attained by laws with at most three support points in $[0,1]$. The quadratic dual extrema are also attained, with}
\begin{align*}
L_n(\mu,\nu)
&=\max\{a+b\mu+c\nu: a+bq+cq^2\le P_n(q)\ \forall q\in[0,1]\},\\
U_n(\mu,\nu)
&=\min\{a+b\mu+c\nu: a+bq+cq^2\ge P_n(q)\ \forall q\in[0,1]\}.
\end{align*}
\end{theorem}

The proof is in Appendix~\ref{app:proof-general-odd-main}.

Compactness of $\cM(\mu,\nu)$ and continuity of $P_n$ give an optimizer, which linearity allows us to choose at an extreme point. \textcolor{black}{The three constraints fix total mass, mean, and second moment.} If a finite-support law has positive weights on four distinct points $q_j$, the four vectors $(1,q_j,q_j^2)$ are linearly dependent. One can perturb the weights in either direction while preserving every constraint, so that law is not extreme. The finite-moment extreme-point theorem extends this argument to general measures \citep{Winkler1988,Pinelis2016}. \textcolor{black}{These three-point laws attain the bounds.}

A quadratic upper envelope $r(q)=a+bq+cq^2\ge P_n(q)$ has known expectation $a+b\mu+c\nu$. A supporting hyperplane of the convex hull of $(q,q^2,P_n(q))$ gives a matching envelope at an interior moment pair. \textcolor{black}{The nonnegative gap $r(q)-P_n(q)$ has zero expectation at an optimizer,} so every positive-weight support point satisfies $r(q)=P_n(q)$. Reversing the inequality gives the lower bound. The global envelope and an attaining distribution together establish sharpness.

\textcolor{black}{\textcolor{black}{This dual representation also gives a direct computation from the moments and the majority polynomial.} For an upper bound, minimize $a+b\mu+c\nu$ over the three coefficients of $r(q)=a+bq+cq^2$, subject to $r(q)\ge P_n(q)$ on $[0,1]$. An exchange method starts with constraints at a finite set of points and solves the resulting linear program. It then checks the polynomial slack $r-P_n$ at both endpoints and every real stationary point in the interval, where its minimum must occur. Any violating point is added to the constraints before solving again. Applying the same procedure to $-P_n$ gives the lower bound. The implementation checks residual slack and adjusts the constant term outward to account for numerical error; Appendix~\ref{app:numerics-main} gives the computation details. \textcolor{black}{Increasing the budget raises the polynomial degree; the dual retains three coefficients.} }

\textcolor{black}{The same moment class also gives sharp bounds on gains between budgets: replacing $P_n$ by $P_n-P_m$ in Theorem~\ref{thm:general-odd-main} optimizes the increment under one latent law. Extremizing laws can change with the budget, so direct increment bounds can be tighter than differences of separately optimized endpoints. For example, the gain from three to five calls is $V_2-V_1=\E[P_2(Q)-P_1(Q)]$. A positive lower bound establishes improvement for every compatible population; a negative upper bound establishes a loss. This uses the same two moments, with the polynomial-objective extension proved in Appendix~\ref{app:proof-general-odd-main}.}

\subsection{Closed-form three-vote bounds}

For three votes, the bounds give an explicit answer to the first useful budget increase beyond one call. Since $P_1(q)=3q^2-2q^3$, the only unobserved quantity is the third moment. Its exact range under $(\mu,\nu)$ yields the following closed form.

\begin{theorem}
\label{thm:three-vote-main}
Let $\mu\in(0,1)$ and $\rho\in[0,1]$, with $\nu=\mu^2+\rho\mu(1-\mu)$. Put $s=\mu(1-\mu)(1-\rho)$. Then, with the usual endpoint conventions at $\mu=0$ and $\mu=1$,
\begin{align}
L_1(\mu,\nu)
&=\nu+2\frac{(\mu-\nu)^2}{1-\mu}
=\mu+s\{2\mu(1-\rho)-1\},
\label{eq:L1-three-main}\\
U_1(\mu,\nu)
&=3\nu-2\frac{\nu^2}{\mu}
=\mu+s\{2\mu+2\rho(1-\mu)-1\}.
\label{eq:U1-three-main}
\end{align}
Moreover $U_1(\mu,\nu)-L_1(\mu,\nu)=2\mu(1-\mu)\rho(1-\rho)\le1/8$, and if $\mu>1/2$, then $V_1>\mu$ for every law in $\cM(\mu,\nu)$ exactly when $\rho<1-1/(2\mu)$.
\end{theorem}

\paragraph{Proof idea: bound the missing third moment.}
Put $m_3=\E Q^3$, $a=\nu/\mu$, and $b=(\mu-\nu)/(1-\mu)$; feasibility places $a,b$ in $[0,1]$. \textcolor{black}{Integrating $q(q-a)^2\ge0$ and $(1-q)(q-b)^2\ge0$ gives}
\begin{equation*}
\frac{\nu^2}{\mu}\ \le\ m_3\ \le\ \nu-\frac{(\mu-\nu)^2}{1-\mu}.
\end{equation*}
Substituting into $V_1=3\nu-2m_3$ gives the two accuracy bounds. Equality is realized on the zero sets of these polynomials: support $\{0,a\}$ attains the upper accuracy, and $\{b,1\}$ attains the lower accuracy, with weights fixed by the mean. Thus the same elementary inequalities both bound the missing moment and locate the populations proving sharpness.

The improvement condition has a direct interpretation in the observable pair table. For $0<\mu<1$,
\[
L_1(\mu,\nu)>\mu
\quad\Longleftrightarrow\quad
\PP(B_1\ne B_2)>2\,\PP(B_1=B_2=0).
\]
Exactly-one-correct pairs must be more than twice as likely as both-wrong pairs. This balance guarantees improvement for every latent law compatible with the two moments. The condition concerns population probabilities; Section~\ref{sec:finite-sample-main} accounts for their estimation from a finite pair table. The proof and extremizers are in Appendix~\ref{app:proof-three-vote-main}.

\subsection{Sharp infinite-vote endpoint}
\label{sec:infinite-main}

The infinite-vote endpoint describes where repetition leads under a fixed latent law. On an example with $q>1/2$, an increasingly large majority becomes correct with probability tending to one; for $q<1/2$, it becomes incorrect with probability tending to one. For each fixed latent law, bounded convergence therefore gives the infinite-vote endpoint
\begin{equation}
\Vinf(\pi):=\lim_{n\to\infty}V_n(\pi)=\pi((1/2,1])+\frac12\pi(\{1/2\}).
\label{eq:main-vinf}
\end{equation}
The half-weight at $1/2$ follows from odd-vote symmetry. Unlike a finite majority polynomial, this objective depends discontinuously on the position of latent mass relative to the voting threshold. Nevertheless, the same two moments admit an exact endpoint characterization.

\begin{theorem}
\label{thm:vinf-main}
Let $U_\infty(\mu,\nu)=\sup_{\pi\in\cM(\mu,\nu)}\Vinf(\pi)$ and $L_\infty(\mu,\nu)=\inf_{\pi\in\cM(\mu,\nu)}\Vinf(\pi)$. Set $G(1/2,1/4)=1/2$, and for every other feasible $(\mu,\nu)$ define
\begin{equation*}
G(\mu,\nu)=
\begin{cases}
\dfrac{\nu-\mu^2}{\nu+1/4-\mu}, & \mu\le 1/2,\ \nu\le \mu/2,\\[8pt]
\min\{1,\,3\mu-2\nu\}, & \text{otherwise}.
\end{cases}
\end{equation*}
Then $U_\infty(\mu,\nu)=G(\mu,\nu)$ and $L_\infty(\mu,\nu)=1-G(1-\mu,\,1-2\mu+\nu)$. These are sharp infima and suprema; one-sided moment-matching sequences may be needed in place of attaining laws.
\end{theorem}

The two branches describe how much population mass can lie on the winning side of $q=1/2$. Writing $\sigma^2=\nu-\mu^2$, the fractional branch is $\sigma^2/\{\sigma^2+(1/2-\mu)^2\}$. When the mean is below the threshold, placing mass just above it requires compensating mass below the mean. The ratio compares the available latent variance with the squared distance that must be crossed. The condition $\nu\le\mu/2$ is precisely what keeps that compensating support point nonnegative.

Outside this regime, the support constraint matters differently. A compensating point cannot move below zero; in the nonsaturated branch, extremal sequences instead use the endpoints $0$ and $1$ together with a point just above $1/2$. The quadratic majorant $3q-2q^2$ matches this geometry and gives $3\mu-2\nu$. When the moments permit all mass to lie above the voting threshold, or to approach such a configuration, the upper endpoint reaches one, explaining the cap. Reflection $Q\mapsto1-Q$ gives the lower endpoint. \textcolor{black}{Appendix~\ref{app:proof-vinf-main} gives the constructions and one-sided limits.}

\begin{figure}[H]
    \centering
    \includegraphics[width=0.86\linewidth]{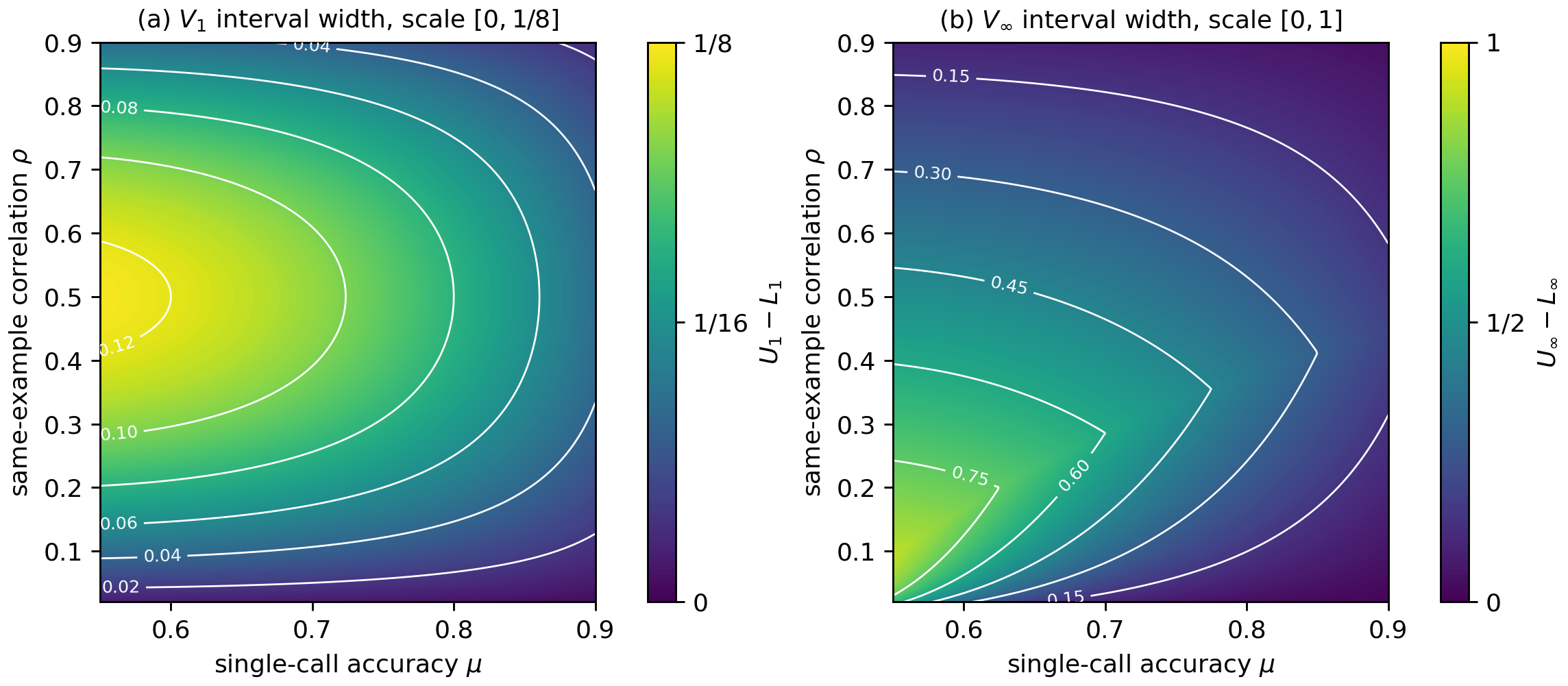}
    \caption{Population identification widths from two moments. (a) Exact three-vote width $U_1-L_1$, on the scale $[0,1/8]$. (b) Exact infinite-vote width $U_\infty-L_\infty$, on the different scale $[0,1]$. Both panels use horizontal $\mu$ and vertical $\rho$; \textcolor{black}{each panel has its own color scale. Widths describe variation over latent laws with known moments.}}
    \label{fig:widths-main}
\end{figure}

Figure~\ref{fig:widths-main} shows the contrast with three votes. At $\mu=1/2$ and $0<\rho<1$, the exact widths are
\begin{equation*}
U_1-L_1=\tfrac12\rho(1-\rho),\qquad
U_\infty-L_\infty=1-\rho.
\end{equation*}
As positive $\rho$ tends to zero, the finite width vanishes while the endpoint width approaches one. At $\rho=0$, $Q=1/2$ is forced and both widths are zero. Thus tight finite-budget bounds do not imply a tight infinite-vote extrapolation: the endpoint has its own sharp geometry.

\subsection{\textcolor{black}{Statistical inference from two calls}}
\label{sec:finite-sample-main}

The population bounds treat $(\mu,\nu)$ as known. The two-call moment estimates are subject to sampling uncertainty. From $N$ labeled example draws, form
\[
X_i=\bigl((B_{i1}+B_{i2})/2,\,B_{i1}B_{i2}\bigr),\qquad
\hat\theta=(\hat\mu,\hat\nu)=N^{-1}\sum_iX_i,
\]
\textcolor{black}{The coordinates record average correctness and joint success. Let $\hat\Sigma$ be their sample covariance.}

For iid draws with nonsingular covariance, use the feasible Wald region
\begin{equation}
\mathcal C_\alpha^{\mathrm W}
=\{\theta\in\mathcal F:
N(\theta-\hat\theta)^\top\hat\Sigma^{-1}(\theta-\hat\theta)
\le\chi^2_{2,1-\alpha}\},
\label{eq:wald-region-main}
\end{equation}
where $\mathcal F=\{(\mu,\nu):0\le\mu\le1,\ \mu^2\le\nu\le\mu\}$. It uses the covariance between the estimated moments and has nominal confidence $1-\alpha$ under the usual large-sample approximation.

\textcolor{black}{For each moment pair in this region, compute its sharp interval. Taking the union gives confidence sets for accuracy and gain over one call:}
\begin{align*}
\mathcal S_{n,\alpha}^{\mathrm W}
&=\bigcup_{(\mu,\nu)\in\mathcal C_\alpha^{\mathrm W}}
[L_n(\mu,\nu),U_n(\mu,\nu)],
\\
\mathcal G_{n,\alpha}^{\mathrm W}
&=\bigcup_{(\mu,\nu)\in\mathcal C_\alpha^{\mathrm W}}
[L_n(\mu,\nu)-\mu,U_n(\mu,\nu)-\mu].
\end{align*}
We report \textcolor{black}{the smallest interval containing each projected set}. Whenever the moment region contains the true pair, these sets cover all corresponding vote accuracies and gains simultaneously for that protocol. The same construction applies to the infinite-vote endpoints in Section~\ref{sec:infinite-main}. The statistical coverage is asymptotic; sharpness refers to the population moment bounds.

\textcolor{black}{Projecting the direct incremental-gain bounds over the same moment region gives simultaneous comparisons between budgets, including comparisons selected after inspecting the intervals. A positive lower gain bound establishes improvement at the nominal confidence level. \textcolor{black}{Appendix~\ref{app:gain-comparisons-main} details gain projection, protocol adjustments, and budget selection.}}

\section{\textcolor{black}{Parametric curve completions}}
\label{sec:completion-main}

\textcolor{black}{A curve completion selects one success-probability distribution matching the moments and uses it to forecast accuracy across budgets. Maximum entropy and a latent-difficulty Gaussian-probit model each give a unique fit for every interior moment pair, with predictions inside the sharp bounds.}

\subsection{\textcolor{black}{Maximum-entropy completion}}

\textcolor{black}{Maximum entropy (ME) selects the density with the largest differential entropy subject to the observed moments \citep{Jaynes1957}. Fixing a mean and variance on the real line gives a Gaussian; on $[0,1]$, the corresponding density is quadratic-exponential:}
{\color{black}
\begin{align*}
\psi(\lambda,\kappa)&=\log\int_0^1\exp\{\lambda(q-1/2)+\kappa(q-1/2)^2\}\,\dd q,\\
f_{\lambda,\kappa}(q)&=\exp\{\lambda(q-1/2)+\kappa(q-1/2)^2-\psi(\lambda,\kappa)\}\1_{[0,1]}(q).
\end{align*}
}
\textcolor{black}{\textcolor{black}{For interior $(\mu,\nu)$, the unique parameters minimize the strictly convex objective}}
{\color{black}
\[
J_{\mu,\nu}(\lambda,\kappa)=\psi(\lambda,\kappa)-\lambda(\mu-1/2)-\kappa(\nu-\mu+1/4).
\]
}
\textcolor{black}{The parameter $\lambda$ tilts the density across the voting threshold, with $\operatorname{sign}(\lambda)=\operatorname{sign}(\mu-1/2)$; at fixed $\lambda$, $\kappa$ shifts mass toward the center or endpoints. The completed curve and its limit are}
{\color{black}
\[
V_n^{\ME}=\int_0^1P_n(q)f_{\lambda,\kappa}(q)\,\dd q,
\qquad V_\infty^{\ME}=\int_{1/2}^1f_{\lambda,\kappa}(q)\,\dd q.
\]
}
\textcolor{black}{Proposition~\ref{prop:maxent-main} gives the entropy characterization and uniqueness; the proof is in Appendix~\ref{app:proofs-main}.}

\subsection{\textcolor{black}{Latent-difficulty Gaussian-probit completion}}

\textcolor{black}{The latent-difficulty Gaussian-probit (LDGP) completion models variation in success probabilities through example difficulty, following normal-ogive item-response models \citep{Albert1992,Lord1980}:}
{\color{black}
\[
Q=\Phi(\eta-\gamma Z),\qquad Z\sim N(0,1),\qquad\gamma\ge0.
\]
}
\textcolor{black}{Here $Z$ represents example difficulty, $\eta$ locates performance relative to this scale, $\gamma$ controls the effect of difficulty on success probability, and $\Phi$ is the standard normal cdf. Writing $r=\gamma^2/(1+\gamma^2)$, moment matching gives}
{\color{black}
\[
t=\Phi^{-1}(\mu),\quad\nu=\Phi_2(t,t;r),\quad
\gamma=\sqrt{\frac{r}{1-r}},\quad\eta=\frac{t}{\sqrt{1-r}},
\]
}
\textcolor{black}{where $\Phi_2(\cdot,\cdot;r)$ is the bivariate standard normal cdf; $r$ is distinct from the correctness correlation $\rho$. For fixed $t$, the second-moment map increases strictly from $\mu^2$ to $\mu$ as $r$ runs from zero to one, giving a unique fit for every interior moment pair. Its vote curve is}
{\color{black}
\[
V_n^{\mathrm{LDGP}}=\E\bigl[P_n\{\Phi(\eta-\gamma Z)\}\bigr],
\qquad V_\infty^{\mathrm{LDGP}}=\Phi(\eta/\gamma).
\]
}
\textcolor{black}{Proposition~\ref{prop:ldgp-main} includes the boundary case $\gamma=0$. Thus two calls determine a full curve under a familiar difficulty model. Table~\ref{tab:completion-curves-main} compares the two completions at matched moments; their finite-budget forecasts are close in the examples shown, while the limiting forecasts reflect different allocations of mass around $q=1/2$.}

\section{Two-call decisions on LLM validation data}
\label{sec:experiments-main}

\begin{figure}[htbp]
\centering
\includegraphics[width=0.96\linewidth]{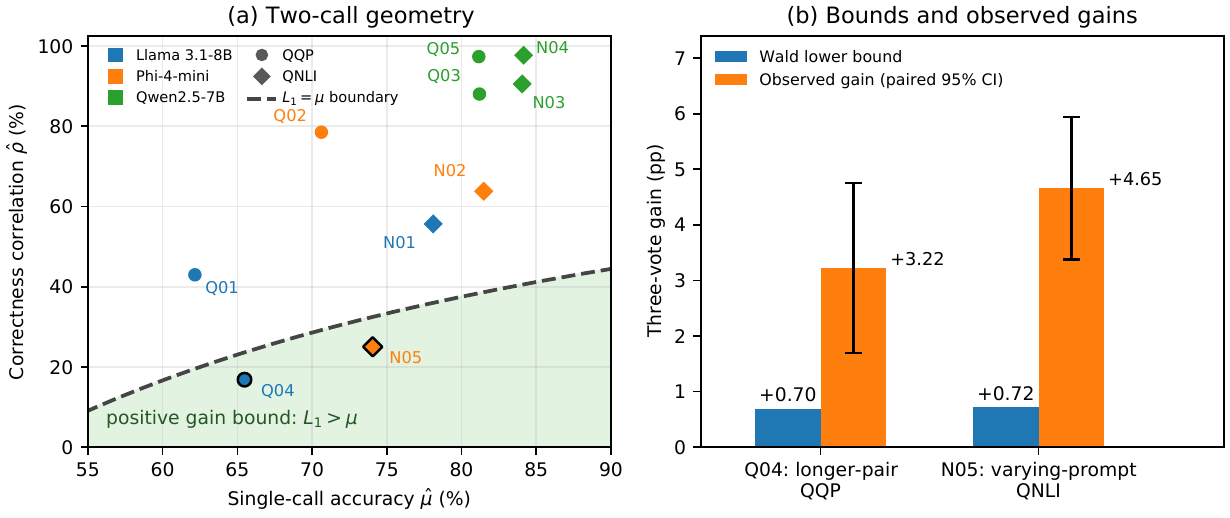}
\caption{\textcolor{black}{Two-call geometry and three-vote gains. (a) Ten settings by single-call accuracy and correctness correlation; colors give the model, markers the task, and outlined markers the held-out settings Q04 and N05. \textcolor{black}{The gray dashed curve bounds the green region where the lower gain bound is positive.} (b) Lower confidence bounds on the three-vote gain for Q04 and N05 (Wald projection, nominal joint confidence $95\%$) and the gains observed with a third call on $1{,}024$ prespecified draws (paired $95\%$ intervals), in percentage points.}}
\label{fig:mu-rho}
\end{figure}

\textcolor{black}{We evaluate binary judgments on QQP (are two questions duplicates?) and QNLI (does a sentence answer a question?) \citep{Wang2019} with Llama 3.1-8B, Phi-4-mini, and Qwen2.5-7B served through Ollama; each call asks a Yes/No question that is scored against the task label. \textcolor{black}{Figure~\ref{fig:mu-rho}a shows ten settings, with Q denoting QQP and N denoting QNLI.} \textcolor{black}{Q01--Q03 and N01--N03 use temperature $1$; Q05 and N04 use Qwen at $0.2$, all on $8{,}192$ examples per task.} }

\textcolor{black}{\textcolor{black}{Two additional protocols, fixed on development data and evaluated on separate held-out examples, assess the gain bound:} Q04 asks Llama at temperature $2$ about longer QQP pairs, and N05 asks Phi at temperature $2$ QNLI questions whose wording is resampled on every call, each with $6{,}144$ draws and a prespecified $1{,}024$-draw subset that receives a third call. Appendix~\ref{app:llm-table} lists all settings and Appendix~\ref{app:positive-protocols-main} the prompts and sampling rules. A gain is the difference of percentage accuracies $100(V_n-V_0)$ within a setting.}

\textcolor{black}{The full direct-prompt grid contains 18 task--model--temperature settings, with five responses per example. We estimate moments from the first two responses. Empirical three-vote accuracy averages over the ten possible triples of recorded responses; five-vote accuracy uses all five. Each gain uses the single-call baseline averaged over the same recorded responses (Appendix~\ref{app:numerics-main}). All empirical three- and five-vote accuracies fall within the projected $95\%$ regions for their settings (Table~\ref{tab:app-all-results}). These are compatibility checks using shared responses.}

\textcolor{black}{The vertical coordinate in Figure~\ref{fig:mu-rho}a, the correlation between two calls on the same example, is what single-call accuracy omits: a population of always-correct and always-wrong examples matches any mean and gains nothing from voting. \textcolor{black}{At a fixed single-call accuracy, higher correctness correlation means fewer disagreements, limiting how much three-vote accuracy can change (Section~\ref{sec:original-conditions-main}). The held-out settings lie inside the shaded region, where every population compatible with the estimated moments gains from three votes. The projected bounds below account for estimation uncertainty. Several direct-prompt settings outside this region show realized gains. Their estimated moments are compatible with both positive and negative population gains.}} \textcolor{black}{On the same QQP examples, raising Llama's temperature from $0.2$ to $2.0$ lowers single-call accuracy and correctness correlation while increasing the five-vote gain from $-0.06$ to $+2.57$ percentage points (Table~\ref{tab:app-all-results}, Q06 and Q07).}

\textcolor{black}{For Q04 and N05, projecting the Wald region of the two-call statistics through the sharp lower bound establishes three-vote gains of at least $+0.70$ and $+0.72$ percentage points at nominal joint confidence $95\%$ (Figure~\ref{fig:mu-rho}b; Appendix~\ref{app:pair-confidence-main}). The third calls then show observed gains of $+3.22$ and $+4.65$ points, with paired $95\%$ intervals $[1.70,4.75]$ and $[3.38,5.93]$ (Table~\ref{tab:confirmation-checks}). \textcolor{black}{The projected minima bound population gains; the third-call estimates measure observed gains.}}

\textcolor{black}{Both gain bounds exceed half a percentage point at the reported nominal joint confidence level. Minimization over the moment region and all compatible populations accounts for sampling uncertainty and the unknown success-probability distribution (Appendix~\ref{app:gain-comparisons-main}).}

\subsection{Budget comparisons}
\label{sec:original-conditions-main}

\textcolor{black}{\textcolor{black}{Two-call upper bounds also quantify how much improvement remains possible at a specified budget.} Since $P_1(q)-q=q(1-q)(2q-1)$, averaging over the population gives}
{\color{black}
\[
|V_1-\mu|\le \E[Q(1-Q)]=\mu-\nu
=\tfrac12\PP(B_1\ne B_2),
\]
}
\textcolor{black}{so three-vote accuracy moves by at most half the disagreement rate. In N04, Qwen at temperature $0.2$, the two calls disagree on about $0.6\%$ of examples; a Wald upper bound on this rate, applied to the inequality and its five-vote analogue, limits the gain to $+0.46$ and $+0.47$ points at familywise $95\%$ confidence across the direct-prompt settings (Appendix~\ref{app:headroom-results-main}). These bounds quantify the limited benefit of three or five calls without evaluating either budget directly.}

\textcolor{black}{The broader direct-prompt grid shows the same pattern: Qwen has the highest single-call accuracy in both tasks, yet its observed five-vote gain is at most $0.17$ percentage points on QNLI and $0.04$ on QQP across the three temperatures. Together with the Llama temperature comparison, this shows why baseline accuracy and repeatability provide different information about the value of extra calls.}

\textcolor{black}{For the same half-percentage-point target, the lower bounds for Q04 and N05 support additional calls, whereas the three- and five-vote upper bounds for N04 exclude that improvement at their reported confidence levels. These comparisons concern gains over each setting's own baseline; absolute accuracy and call cost also matter for deployment.}

The choice of call budget also matters because single-call rankings can change after voting. \textcolor{black}{Among the additional QNLI temperature settings}, N09 (Phi-4-mini at $T=2.0$) trails N06 (Llama 3.1-8B at $T=0.2$) by $0.86$ with one call but leads by $1.04$ after five votes, measured as differences of percentage accuracies. This voting-induced rank reversal uses recorded-repeat comparisons on the same examples (Table~\ref{tab:overtaking-main}). It shows why protocols should be compared at the intended vote budget rather than ranked by single-call accuracy alone.

\textcolor{black}{The point forecasts also connect the two-call moments to empirical voting accuracy. Across the 18 direct-prompt settings, LDGP has mean absolute errors of $0.21$ and $0.22$ percentage points at three and five votes, compared with $0.36$ and $0.35$ for the interval midpoint (Table~\ref{tab:point-completion-main}). Both model completions overpredict on average. Appendix~\ref{app:mixture-main} also reports randomized-mixture results based on fitted per-example probabilities, illustrating how independent component selection can change repeatability and voting gains.}

The supplementary nine-call GSM8K evaluation records positive strict-majority correctness gains but leaves their sign unresolved by the two-call Wald intervals. Appendix~\ref{app:gsm8k-details-main} reports the full curve and separate plurality outcomes; Appendix~\ref{app:dependence-stress-main} examines the role of conditional independence.

\section{Conclusion}
\label{sec:discussion-main}

Two labeled calls per example turn repeatability into evidence for allocating test-time compute. \textcolor{black}{Their joint correctness table identifies the moments needed for sharp, computable accuracy bounds at every finite budget and for gains between budgets. The three-vote case yields a direct improvement criterion.} \textcolor{black}{Projecting a confidence region for the two moments through these bounds supports budget comparisons with validation uncertainty included.} In the binary-judgment evaluations, recorded gains vary across settings; \textcolor{black}{the two-call bounds establish positive gains in some settings and limit the possible improvement in another.}

The infinite-vote bounds extend this account to limiting accuracy, while the moment-matched completions turn the same information into full-curve forecasts. The observed rank reversal further shows why model choice and call budget should be considered together. \textcolor{black}{A contextual extension could use example features to estimate conditional moments and guide individual call budgets, with regression uncertainty propagated through the bounds. The same two-call validation sample supports accuracy and gain bounds across finite deployment budgets.}

\clearpage
\appendix
\section{Proofs of main results}
\label{app:proofs-main}

This appendix gives formal proofs for the statements whose main-text presentations are abbreviated.

\subsection{Even-budget collapse}\label{app:even-collapse-main}

Let $X\sim\Bin(2n-1,q)$ be the number of correct responses among the first $2n-1$ calls and let $Y\sim\Bern(q)$ be the last call, independent of $X$. For $2n$ binary votes with uniform random tie-breaking, the success probability is
\[
\PP(X+Y>n)+\frac12\PP(X+Y=n).
\]
Compared with the preceding odd-budget accuracy $\PP(X\ge n)$, only the boundary events $X=n-1$ and $X=n$ matter. Since
\[
q\PP(X=n-1)=(1-q)\PP(X=n),
\]
which follows from $\binom{2n-1}{n-1}=\binom{2n-1}{n}$, the two boundary contributions cancel and this $2n$-call rule has the same success probability as the $(2n-1)$-call strict-majority rule.

\subsection{Proof of Proposition~\ref{prop:moments-main}}\label{app:proof-moments-main}

Conditional on $q_i$, the two correctness indicators are independent Bernoulli variables with success probability $q_i$. Therefore
\[
\E[B_{i1}\mid q_i]=q_i,\qquad
\E[B_{i1}B_{i2}\mid q_i]=\E[B_{i1}\mid q_i]\E[B_{i2}\mid q_i]=q_i^2.
\]
Averaging over examples gives $\E B_{i1}=\mu$ and $\E(B_{i1}B_{i2})=\nu$. Since $B_{ij}$ is binary,
\[
\Var(B_{ij})=\mu(1-\mu),\qquad
\Cov(B_{i1},B_{i2})=\nu-\mu^2,
\]
which yields \eqref{eq:rho-main} whenever $0<\mu<1$.

The feasibility inequalities follow from $0\le Q\le1$: Jensen's inequality gives $\nu=\E Q^2\ge(\E Q)^2=\mu^2$, and $Q^2\le Q$ gives $\nu\le\mu$. Conversely, every pair satisfying $\mu^2\le\nu\le\mu$ is attainable. If $\mu=0$ or $\mu=1$, the only possible laws are $\delta_0$ or $\delta_1$. For $0<\mu<1$, the two-point law
\[
\pi=(1-p)\delta_0+p\delta_a,\qquad
a=\frac{\nu}{\mu},\qquad
p=\frac{\mu^2}{\nu},
\]
with the limiting interpretation when $\nu=\mu^2$, has mean $\mu$ and second moment $\nu$ because $a\in[\mu,1]$ and $p\in[0,1]$. This proves exact feasibility. Finally, \eqref{eq:pair-probs-main} follows from
\[
\PP(B_{i1}=1,B_{i2}=1)=\E Q^2,\qquad
\PP(B_{i1}=1,B_{i2}=0)=\PP(B_{i1}=0,B_{i2}=1)=\E[Q(1-Q)].
\]

\subsection{Mean-only examples and the one-call envelope}\label{app:mean-only-main}

For $\pi^{\mathrm{det}}_\mu=(1-\mu)\delta_0+\mu\delta_1$, the majority polynomial satisfies $P_n(0)=0$ and $P_n(1)=1$, hence $V_n(\pi^{\mathrm{det}}_\mu)=\mu$ for every odd budget. For $\pi^{\mathrm{hom}}_\mu=\delta_\mu$, $V_n(\pi^{\mathrm{hom}}_\mu)=P_n(\mu)$ by definition. The endpoint formula follows by evaluating \eqref{eq:main-vinf} for these two laws.

For three votes, the mean-only lower and upper ranges are
\begin{equation*}
L_1^{\mathrm{one}}(\mu)=
\begin{cases}
P_1(\mu), & \mu\le 1/4,\\[2pt]
(9\mu-1)/8, & \mu\ge 1/4,
\end{cases}
\qquad
U_1^{\mathrm{one}}(\mu)=
\begin{cases}
9\mu/8, & \mu\le 3/4,\\[2pt]
P_1(\mu), & \mu\ge 3/4.
\end{cases}
\end{equation*}
Indeed,
\[
P_1(q)=3q^2-2q^3,\qquad P_1'(q)=6q(1-q),
\]
so the lower convex envelope of $P_1$ on $[0,1]$ is $P_1(q)$ on $[0,1/4]$ and the tangent chord
\[
\ell_-(q)=\frac{9q-1}{8}
\]
on $[1/4,1]$. The two pieces match in value and slope at $q=1/4$, and $\ell_-(1)=P_1(1)=1$. Jensen's inequality applied to the lower convex envelope gives the lower bound; the linear branch is attained by laws supported on $\{1/4,1\}$. By the symmetry $P_1(q)=1-P_1(1-q)$, the upper concave envelope is $\ell_+(q)=9q/8$ on $[0,3/4]$ and $P_1(q)$ on $[3/4,1]$; the linear branch is attained by laws supported on $\{0,3/4\}$. This proves the upper formula as well. Since $L_1^{\mathrm{one}}(\mu)\le\mu$ for every interior $\mu$, one labeled call alone cannot guarantee three-vote improvement.

\subsection{Proof of Theorem~\ref{thm:general-odd-main}}\label{app:proof-general-odd-main}

The moment class \eqref{eq:momentclass-main} is tight and weakly closed on the compact space $[0,1]$, hence compact. Because $P_n$ is continuous, $\pi\mapsto\int P_n\,d\pi$ is continuous and linear, so both primal extrema are attained.

An extremizer may be chosen to have at most three support points. The feasible set is a two-moment slice of the probability simplex, and a linear functional attains its maximum and minimum at extreme points of this compact convex slice. Under the three linear constraints $\int1\,d\pi=1$, $\int q\,d\pi=\mu$, and $\int q^2\,d\pi=\nu$, every extreme point has support size at most three. In the atomic case this is immediate: if four distinct support points $q_1,\ldots,q_4$ carry positive mass, then the four vectors $(1,q_j,q_j^2)\in\mathbb R^3$ are linearly dependent, so a nonzero signed perturbation preserves all three moments; small positive and negative perturbation sizes keep the measure nonnegative, contradicting extremality. \textcolor{black}{The general case follows from the finite-moment extreme-point theorem of Winkler--Caratheodory.}

For the quadratic dual, first consider the upper endpoint at an interior moment pair $\mu^2<\nu<\mu$. Any quadratic $r(q)=a+bq+cq^2$ satisfying $r(q)\ge P_n(q)$ on $[0,1]$ gives
\[
\int P_n(q)\,d\pi(q)\le \int r(q)\,d\pi(q)=a+b\mu+c\nu
\]
for every $\pi\in\cM(\mu,\nu)$, so the dual majorant value is an upper bound. To see exactness and attainment, consider the compact convex set
\[
\mathcal A_n=\operatorname{conv}\{(1,q,q^2,P_n(q)):q\in[0,1]\}\subset\mathbb R^4 .
\]
Its vertical section at $(1,\mu,\nu)$ is the closed interval $\{(1,\mu,\nu,v):L_n(\mu,\nu)\le v\le U_n(\mu,\nu)\}$. At an interior moment pair, a supporting hyperplane at the upper boundary has a nonzero last coefficient: otherwise it would support the projected moment region at an interior point. It can therefore be normalized to have last coefficient $-1$, giving coefficients $(a,b,c)$ such that $a+bq+cq^2\ge P_n(q)$ for every $q$ and $a+b\mu+c\nu=U_n(\mu,\nu)$. Thus the dual minimum is attained and equals the primal upper endpoint. Applying the same argument to the lower boundary, or equivalently to $-P_n$, gives an attained quadratic minorant and the formula for $L_n$.

At a degenerate moment pair, the finite-vote polynomial's bounded second derivative gives matching quadratic envelopes for the unique feasible law. The general continuous-objective extension is restricted to interior moment pairs, omitting degenerate cases such as $(0,0)$; there the same proof gives attained quadratic majorants and minorants and three-atom primal extremizers. In particular, any polynomial objective, or any linear combination of finite-vote accuracies, is optimized in the same three-atom moment space. Quantities such as the largest possible marginal gain $V_3-V_1$ are therefore obtained by running the same program with objective $P_3-P_1$.

\subsection{Proof of Theorem~\ref{thm:three-vote-main}}\label{app:proof-three-vote-main}

For three votes,
\[
P_1(q)=3q^2-2q^3,
\qquad
V_1=3\nu-2m_3,\quad m_3:=\E Q^3.
\]
Thus the upper bound on $V_1$ is obtained by minimizing $m_3$, and the lower bound by maximizing $m_3$, under the first two moments.

For any $a\in[0,1]$,
\[
q(q-a)^2\ge0.
\]
Expanding and integrating gives
\[
m_3-2a\nu+a^2\mu\ge0.
\]
Choosing $a=\nu/\mu$ for $0<\mu<1$ yields
\[
m_3\ge\frac{\nu^2}{\mu}.
\]
Equality is attained by a law supported on $\{0,\nu/\mu\}$ with the appropriate weights, so
\[
U_1(\mu,\nu)=3\nu-2\frac{\nu^2}{\mu}.
\]

Similarly,
\[
(1-q)(q-b)^2\ge0
\]
implies
\[
\nu-2b\mu+b^2-m_3+2b\nu-b^2\mu\ge0.
\]
Choosing $b=(\mu-\nu)/(1-\mu)$ gives
\[
m_3\le \nu-\frac{(\mu-\nu)^2}{1-\mu}.
\]
Equality is attained by a law supported on $\{(\mu-\nu)/(1-\mu),1\}$, hence
\[
L_1(\mu,\nu)=3\nu-2\left\{\nu-\frac{(\mu-\nu)^2}{1-\mu}\right\}
=\nu+2\frac{(\mu-\nu)^2}{1-\mu}.
\]
The endpoint cases $\mu=0$ and $\mu=1$ are forced by feasibility.

Substituting $\nu=\mu^2+\rho\mu(1-\mu)$ gives the equivalent $\rho$-form in \eqref{eq:L1-three-main}--\eqref{eq:U1-three-main}. Direct subtraction yields
\[
U_1-L_1=2\mu(1-\mu)\rho(1-\rho)\le 2\cdot\frac14\cdot\frac14=\frac18.
\]
Finally, for $\mu>1/2$, uniform improvement over one call is equivalent to $L_1(\mu,\nu)>\mu$. By \eqref{eq:L1-three-main},
\[
L_1-\mu=\mu(1-\mu)(1-\rho)\{2\mu(1-\rho)-1\}.
\]
The prefactor is positive except in the degenerate deterministic case, so strict improvement holds exactly when $2\mu(1-\rho)>1$, i.e.,
\[
\rho<1-\frac{1}{2\mu}.
\]

Figure~\ref{fig:gain-map-main} visualizes the three-vote gain lower bound $L_1-\mu$ over the feasible two-call moment region. The zero contour is exactly the theorem's improvement boundary $\rho=1-1/(2\mu)$ for $\mu>1/2$.

\begin{center}
    \includegraphics[width=0.68\linewidth]{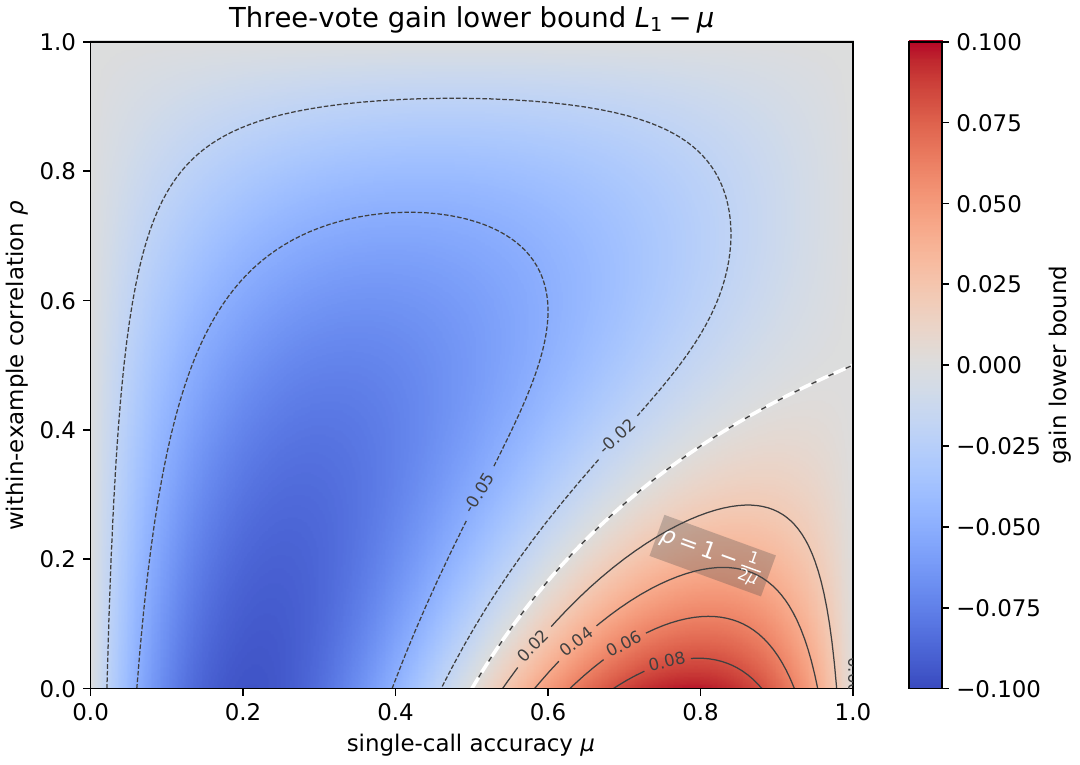}
    \captionof{figure}{Three-vote gain lower bound $L_1-\mu$ implied by Theorem~\ref{thm:three-vote-main}. Warm regions guarantee a uniform gain from three votes over one call; cool regions indicate that no such uniform gain is guaranteed from the two-call moments alone. The dashed white contour is the zero-gain boundary $\rho=1-1/(2\mu)$.}
    \label{fig:gain-map-main}
\end{center}

\subsection{Proof of Theorem~\ref{thm:vinf-main}}\label{app:proof-vinf-main}\label{app:infinite-endpoint-main}

Let
\[
h(q)=\1\{q>1/2\}+\frac12\1\{q=1/2\}.
\]
The upper endpoint is the supremum of $\int h\,d\pi$ over $\cM(\mu,\nu)$. \textcolor{black}{The deterministic cases $\mu=0,1$ are immediate.} The exceptional point $(\mu,\nu)=(1/2,1/4)$ forces $Q\equiv1/2$, so the value is $1/2$.

For the first nontrivial upper branch, assume $\mu\le1/2$ and $\nu\le\mu/2$. For $a<1/2$ define
\[
\psi_a(q)=\left(\frac{q-a}{1/2-a}\right)^2 .
\]
Then $\psi_a(q)\ge h(q)$ on $[0,1]$: it is nonnegative below the threshold, equals one at the threshold, and is at least one above it. Therefore
\[
\int h\,d\pi
\le
\frac{\nu-2a\mu+a^2}{(1/2-a)^2}.
\]
Minimizing the right side over $a<1/2$ gives
\[
a^\star=\frac{\mu/2-\nu}{1/2-\mu}
\]
and the value
\[
\frac{\nu-\mu^2}{\nu+1/4-\mu}.
\]
Here the denominator is positive away from the exceptional point because
\[
\nu+\frac14-\mu=(\nu-\mu^2)+(\mu-1/2)^2.
\]
On this branch, feasibility gives $a^\star\le\mu<1/2$, while $\nu\le\mu/2$ is equivalent to $a^\star\ge0$. Thus the optimizing quadratic is admissible. If $\nu=\mu^2$, the degenerate law $\delta_\mu$ is sharp. Otherwise, put $t=1/2+\varepsilon$ and
\[
x_\varepsilon=\frac{\mu t-\nu}{t-\mu},\qquad
p_\varepsilon=\frac{\mu-x_\varepsilon}{t-x_\varepsilon}.
\]
For all sufficiently small $\varepsilon>0$, the branch inequalities give $0\le x_\varepsilon<1/2<t$ and $0\le p_\varepsilon\le1$. The two-point law $(1-p_\varepsilon)\delta_{x_\varepsilon}+p_\varepsilon\delta_t$ matches $(\mu,\nu)$, and its endpoint value is $p_\varepsilon$, which tends to $(\nu-\mu^2)/(\nu+1/4-\mu)$ as $\varepsilon\downarrow0$.

For the remaining upper branch, the quadratic
\[
\psi_0(q)=3q-2q^2
\]
satisfies $\psi_0(q)\ge h(q)$ on $[0,1]$: it is nonnegative on $[0,1/2)$, equals one at $1/2$, and is at least one on $(1/2,1]$. Hence
\[
\int h\,d\pi\le 3\mu-2\nu,
\]
with the trivial additional bound $\int h\,d\pi\le1$. If $3\mu-2\nu>1$, a two-point law supported on $\{x,1\}$ with $x=(\mu-\nu)/(1-\mu)>1/2$ attains endpoint value one whenever this branch is active. If $3\mu-2\nu\le1$, set $s=1/2+\varepsilon$ and use support $\{0,s,1\}$ with weights
\[
w_s=\frac{\mu-\nu}{s(1-s)},\qquad
w_1=\frac{\nu-s\mu}{1-s},\qquad
w_0=1-w_s-w_1 .
\]
They match the two moments. The weights are nonnegative for all sufficiently small $\varepsilon>0$. Feasibility gives $w_s\ge0$. On the present branch, either $\mu>1/2$ or $\nu>\mu/2$, and feasibility gives $\nu/\mu>1/2$ when $\mu>0$; hence $w_1\ge0$ after taking $s\in(1/2,\nu/\mu)$. Finally,
\[
w_0=\frac{\nu-\mu+s(1-\mu)}{s}.
\]
If $1-3\mu+2\nu>0$, then $w_0>0$ for $s$ close to $1/2$. On the boundary $1-3\mu+2\nu=0$, this formula gives $w_0=(1-\mu)(s-1/2)/s\ge0$. Since $s>1/2$, the endpoint value is $w_s+w_1=1-w_0$, which tends to $3\mu-2\nu$ as $\varepsilon\downarrow0$. This proves the stated $U_\infty$ formula.

For the lower endpoint, use the exact symmetry
\[
h(q)+h(1-q)=1,
\]
including at $q=1/2$, where both terms equal $1/2$. If $\widetilde Q=1-Q$, then
\[
\E\widetilde Q=1-\mu,\qquad
\E\widetilde Q^2=1-2\mu+\nu.
\]
Therefore
\[
L_\infty(\mu,\nu)
=
1-U_\infty(1-\mu,\,1-2\mu+\nu),
\]
and substituting the upper formula with these reflected moments gives the displayed reflection formula. The same limiting constructions prove sharpness of the lower bound.

\subsection{Proof of Proposition~\ref{prop:maxent-main}}

Let $X=Q-1/2$. The log-partition function
\[
\psi(\lambda,\kappa)=\log\int_{-1/2}^{1/2}e^{\lambda x+\kappa x^2}\,dx
\]
is finite and smooth for all $(\lambda,\kappa)\in\mathbb R^2$. Differentiating under the integral sign gives
\[
\nabla\psi(\lambda,\kappa)=
\bigl(\E_{\lambda,\kappa}X,\E_{\lambda,\kappa}X^2\bigr),
\]
and its Hessian is the covariance matrix of $(X,X^2)$ under the corresponding density. This covariance matrix is positive definite because the curve $x\mapsto(x,x^2)$ is not contained in any affine line on an interval of positive length. Hence $\psi$ is strictly convex.

The convex support of $(X,X^2)$ has interior
\[
\{(m,s):-1/2<m<1/2,\ m^2<s<1/4\}.
\]
Let the target be $(m,s)=(\mu-1/2,\nu-\mu+1/4)$ in this interior. The dual objective $J_{\mu,\nu}(\lambda,\kappa)=\psi(\lambda,\kappa)-\lambda m-\kappa s$ is strictly convex. It is also coercive: along any sequence $(\lambda_j,\kappa_j)$ with norm tending to infinity, a subsequential direction $u$ exposes a strictly positive gap between the support function $\sup_x u\cdot(x,x^2)$ and $u\cdot(m,s)$, because $(m,s)$ is an interior point of the convex support. The compact-support Laplace bound for $\psi$ then makes $J_{\mu,\nu}(\lambda_j,\kappa_j)\to\infty$. Hence $J_{\mu,\nu}$ has a unique minimizer, and its first-order conditions give
\[
\E Q=\mu,\qquad \E Q^2=\nu.
\]

For entropy optimality, let $g$ be any absolutely continuous density on $[0,1]$ with the same two moments as $f_{\lambda,\kappa}$. \textcolor{black}{Since $\log f_{\lambda,\kappa}$ is affine in $1$, $q-1/2$, and $(q-1/2)^2$, matching moments gives}
\[
H(g)=H(f_{\lambda,\kappa})-D_{\mathrm{KL}}(g\,\|\,f_{\lambda,\kappa}).
\]
\textcolor{black}{\textcolor{black}{Nonnegative KL divergence, zero exactly when $g=f_{\lambda,\kappa}$ almost everywhere, proves uniqueness.}}

Finally, for $q\in[0,1]$,
\[
\frac{f_{\lambda,\kappa}(q)}{f_{\lambda,\kappa}(1-q)}
=
\exp\{2\lambda(q-1/2)\}.
\]
The map $\lambda\mapsto\E_{\lambda,\kappa}Q$ is strictly increasing for fixed $\kappa$ because its derivative is $\Var_{\lambda,\kappa}(Q)>0$, and at $\lambda=0$ the density is symmetric around $1/2$. Hence $\operatorname{sign}(\lambda)=\operatorname{sign}(\mu-1/2)$.

\subsection{Proof of Proposition~\ref{prop:ldgp-main}}

Let $Z,E_1,E_2$ be independent standard normal variables. Conditional on $Z$, two independent latent responses from the LDGP model satisfy
\[
Q^2=\PP(E_1\le \eta-\gamma Z,\ E_2\le \eta-\gamma Z\mid Z).
\]
After averaging over $Z$, this is the probability that the two normal variables $E_1+\gamma Z$ and $E_2+\gamma Z$ are both at most $\eta$. After standardization, they are bivariate standard normal with common threshold $t=\eta/\sqrt{1+\gamma^2}$ and correlation $r=\gamma^2/(1+\gamma^2)$. Hence
\[
\E Q=\Phi(t),\qquad \E Q^2=\Phi_2(t,t;r).
\]
For a target $\mu$, $t=\Phi^{-1}(\mu)$ is fixed. As $r$ increases from $0$ to $1$, the bivariate normal cdf $\Phi_2(t,t;r)$ increases continuously from $\Phi(t)^2=\mu^2$ to $\Phi(t)=\mu$. \textcolor{black}{Plackett's formula gives strict monotonicity on $(0,1)$}
\[
\frac{\partial}{\partial r}\Phi_2(t,t;r)=\phi_2(t,t;r)>0.
\]
For an interior feasible $\nu$, there is therefore a unique $r\in(0,1)$ matching $\nu$, and then $\gamma=\sqrt{r/(1-r)}$ and $\eta=t/\sqrt{1-r}$ are unique. The boundary $\nu=\mu^2$ corresponds to $r=0$ and $\gamma=0$, while $\nu=\mu$ is the degenerate limit $r\uparrow1$ and $\gamma\to\infty$ rather than a finite-parameter interior law. The formula for $V_n^{\mathrm{LDGP}}$ is the definition of the completed curve, and the endpoint follows from
\[
\PP\{\Phi(\eta-\gamma Z)>1/2\}=\PP(\eta-\gamma Z>0)=\Phi(\eta/\gamma)
\]
when $\gamma>0$. When $\gamma=0$, $Q\equiv\mu$; applying the definition of $V_\infty$ gives endpoint $1$, $1/2$, or $0$ according as $\mu>1/2$, $\mu=1/2$, or $\mu<1/2$.

\section{Pair-table confidence and held-out confirmation}
\label{app:pair-confidence-main}

\subsection{\textcolor{black}{Gain bounds and budget comparisons}}\label{app:gain-comparisons-main}

For three votes, the lower endpoint of the projected gain is the convex minimization
\begin{equation}
\underline\Delta_{1,\alpha}^{\mathrm W}
=\min_{(\mu,\nu)\in\mathcal C_\alpha^{\mathrm W}}
\left\{\nu-\mu+\frac{2(\mu-\nu)^2}{1-\mu}\right\},
\label{eq:wald-gain-main}
\end{equation}
with the endpoint convention from Theorem~\ref{thm:three-vote-main}. A positive value establishes a gain at the nominal Wald confidence level over the same protocol's single-call accuracy. The moment pair and single-call baseline are optimized together, rather than subtracting separately calibrated accuracy limits. \textcolor{black}{The pair-table covariance and held-out calculations follow below.}

\textcolor{black}{For $J$ fixed protocols, use $\mathcal C_{\alpha/J}^{\mathrm W}$ for each to obtain nominal simultaneous confidence $1-\alpha$ by Bonferroni correction. Within a protocol, the same moment region covers all vote budgets. A positive lower gain bound establishes improvement; the upper bound describes how large that improvement could be. An interval containing zero leaves the sign unresolved.}

\textcolor{black}{One may therefore compare the gain intervals for several budgets using the same two-call validation data. Whenever the moment region contains the true pair, all projected comparisons hold together, including comparisons chosen after inspecting the intervals. This supports budget selection even when accuracy varies nonmonotonically with the number of calls.}

\textcolor{black}{\textcolor{black}{For a required accuracy target, the projected accuracy intervals identify} the smallest evaluated budget whose lower bound exceeds the target. A budget whose upper bound falls below the target can be excluded. When deciding whether to increase an existing budget, the corresponding gain interval instead answers whether the additional calls meet a required improvement. Both decisions use the same two-call validation sample. The budget comparison therefore separates the cost of learning about repetition from the number of calls eventually used at deployment.}

\subsection{Wald gain projection from the pair table}

\textcolor{black}{Let $n_{ab}$ count example draws with first-call score $a$ and second-call score $b$, where $a,b\in\{0,1\}$. Put $N=\sum_{a,b}n_{ab}$ and $D=n_{01}+n_{10}$.} \textcolor{black}{\textcolor{black}{The estimates and covariance are}}
\begin{equation}
\hat\mu=\frac{2n_{11}+D}{2N},\qquad
\hat\nu=\frac{n_{11}}N,\qquad
\hat\Sigma=\frac{N}{N-1}
\left\{
\begin{pmatrix}
(\hat\mu+\hat\nu)/2 & \hat\nu\\
\hat\nu & \hat\nu
\end{pmatrix}
-\hat\theta\hat\theta^\top
\right\}.
\label{eq:pair-covariance-app}
\end{equation}
This follows because $X_i$ takes only the values $(0,0)$, $(1/2,0)$, and $(1,1)$. Thus the pair counts suffice to reconstruct both the moment estimate and its Wald ellipse.

For each of Q04 and N05, we minimize the sharp lower gain in \eqref{eq:wald-gain-main} over $\mathcal C_{0.025}^{\mathrm W}$. The objective is convex: its nonlinear term is a quadratic divided by the positive affine function $1-\mu$. Both fitted covariance matrices are nonsingular, and the minimizing moment pairs lie in the feasible interior. The two per-setting confidence levels of $97.5\%$ give nominal joint confidence $95\%$ by Bonferroni correction, conditional on the fixed development-selected protocols. The projection gives a lower confidence bound on the population gain.

\subsection{Held-out populations and inference protocols}
\label{app:positive-protocols-main}
Setting Q04 is QQP with Llama 3.1-8B; N05 is QNLI with Phi-4-mini. Both protocols were chosen using development data and fixed before confirmation. Each setting uses $6{,}144$ uniform independent draws with replacement from a source pool disjoint from development and the direct-prompt GLUE evaluation. The Q04 pool contains $45{,}406$ QQP pairs with more than 30 lowercased alphanumeric words across the two questions; the N05 pool contains $72{,}414$ QNLI examples. Repeated source examples receive fresh calls. A subset of $1{,}024$ draws, chosen before observing outcomes, receives a third call for the empirical check.

Both models run through Ollama at temperature $2.0$ and top-$p=0.95$, without native thinking \citep{Ollama2026GenerateAPI,Ollama2026Llama31,Ollama2026Phi4Mini}. Q04 uses a direct question about whether the two inputs ask the same thing. N05 uses eight fixed formulations: the direct entailment question has probability $9/16$, and each alternative has probability $1/16$. Formulations are sampled independently on every call, and negative formulations are inverted to the original task label before voting. This defines one fixed randomized protocol; exact templates and remaining generation settings are supplied with the code.

Responses are parsed as binary Yes/No labels and scored against the task label. Invalid or truncated responses map to canonical No regardless of the formulation; all confirmation responses were valid. The sampling unit is the example draw, and the confidence calculation assumes fresh, conditionally independent calls under each fixed protocol.
\subsection{Pair counts and gain bounds}
\label{app:positive-counts-main}
\textcolor{black}{Table~\ref{tab:confirmation-counts} reports the pair counts for Q04 and N05; Table~\ref{tab:confirmation-limits} gives their moment estimates and three-vote gain bounds. $W$ and $C$ denote incorrect and correct scores. The plug-in endpoint uses estimated moments; the Wald lower bound also accounts for sampling uncertainty.}
\begin{table}[H]
\centering\small
\begin{tabular}{lrrrrr}
\toprule
Setting & $N$ & $WW$ & $WC$ & $CW$ & $CC$\\
\midrule
Q04: QQP, direct, length-defined & 6144 & 966 & 1156 & 1154 & 2868\\
N05: QNLI, fixed prompt sampler & 6144 & 709 & 923 & 847 & 3665\\
\bottomrule
\end{tabular}
\caption{First-two-call counts for Q04 and N05, used for the Wald moment regions. The third calls are excluded.}
\label{tab:confirmation-counts}
\end{table}

\begin{table}[H]
\centering\small
\color{black}
\setlength{\tabcolsep}{6pt}
\begin{tabular}{lrrrr}
\toprule
Setting & $\hat\mu$ (\%) & $\hat\rho$ (\%) & \shortstack{Plug-in\\lower gain} & \shortstack{Wald\\lower gain}\\
\midrule
Q04 (QQP) & 65.48 & 16.83 & $+1.68$ & $+0.70$\\
N05 (QNLI) & 74.06 & 25.03 & $+1.59$ & $+0.72$\\
\bottomrule
\end{tabular}
\caption{\textcolor{black}{Three-vote gain lower bounds from the first two calls. Gains are differences of percentage accuracies: the plug-in column is $100\{L_1(\hat\mu,\hat\nu)-\hat\mu\}$ and the Wald column is $100\underline\Delta_{1,0.025}^{\mathrm W}$, at nominal joint confidence $95\%$ across Q04 and N05. Values are rounded after calculation. Table~\ref{tab:confirmation-counts} reports the pair counts.}}
\label{tab:confirmation-limits}
\end{table}

\subsection{Empirical three-call gains and a direct-prompt comparison}
\label{app:positive-checks-main}
For each of Q04 and N05 let $m=1024$, $S_i=B_{i1}+B_{i2}+B_{i3}$ \textcolor{black}{on its three-call subset, and}
\[
Z_i=\1\{S_i\ge2\}-S_i/3,\qquad
\widehat\Delta_{\mathrm{check}}=m^{-1}\sum_i Z_i,\qquad
\widehat{\mathrm{se}}=\operatorname{sd}(Z_1,\ldots,Z_m)/\sqrt m.
\]
The paired Wald $95\%$ interval is $\widehat\Delta_{\mathrm{check}}\pm1.96\,\widehat{\mathrm{se}}$, using the sample standard deviation. Its single-call baseline averages the three calls on the subset. These checks reuse the first two calls used to construct the gain bound and do not independently test its coverage.
\begin{table}[H]
\centering\small
\setlength{\tabcolsep}{5pt}
\begin{tabular}{lrrrrl}
\toprule
Setting & $m$ & Single (\%) & Majority (\%) & Gain & Paired Wald 95\% interval\\
\midrule
Q04 (QQP) & 1024 & 64.65 & 67.87 & $+3.22$ & $[+1.70,+4.75]$\\
N05 (QNLI) & 1024 & 76.50 & 81.15 & $+4.65$ & $[+3.38,+5.93]$\\
\bottomrule
\end{tabular}
\caption{Predesignated three-call checks for Q04 and N05. Gains and their intervals are differences of percentage accuracies, $100\widehat\Delta_{\mathrm{check}}$. The paired Wald intervals summarize observed gains; Table~\ref{tab:confirmation-limits} projects two-call moment ellipses through the sharp lower gain.}
\label{tab:confirmation-checks}
\end{table}

A matched direct-prompt QNLI control achieves $84.28\%$ single-call accuracy on the same $1{,}024$ N05 draws, compared with $81.15\%$ for N05's randomized protocol after three votes. This control is distinct from the direct-prompt QNLI settings, which use a different evaluation pool. The positive lower gain bound establishes improvement over N05's own single-call baseline; it does not establish superiority to direct prompting.

\section{Computation details}
\label{app:numerics-main}

\paragraph{Finite-budget computation.}
Finite-budget endpoints use the quadratic-envelope dual in Theorem~\ref{thm:general-odd-main}. An exchange optimization constructs quadratic envelopes, checking their polynomial slack at both endpoints and at all real stationary points in $[0,1]$. Adjusting the quadratic constant for numerical slack gives an outer endpoint. Wald projection optimizes over the moment ellipse as well as the latent law. Degenerate moment pairs are evaluated from their uniquely determined law.
\paragraph{Wald regions and projection.}
\label{app:moment-regions-main}
The sample covariance is $\hat\Sigma=(N-1)^{-1}\sum_i(X_i-\hat\theta)(X_i-\hat\theta)^\top$, equivalently \eqref{eq:pair-covariance-app}. Every reported two-call confidence interval uses the feasible Wald region \eqref{eq:wald-region-main}. The direct-prompt QQP and QNLI settings and the GSM8K curve use $\alpha=0.05$ per setting; Q04 and N05 use $\alpha=0.025$ each. The low-temperature gain comparison also reports an 18-condition Bonferroni adjustment. The without-replacement GLUE and GSM8K samples use the example-level covariance without a finite-population correction. All confidence levels have a large-sample interpretation.

\begin{table}[H]
\centering
\small
\setlength{\tabcolsep}{7pt}
\begin{tabular}{lrrr}
\toprule
Quantity & Minimum SE & Median SE & Maximum SE \\
\midrule
$\hat\mu$ & 0.38 & 0.42 & 0.51 \\
$\hat\nu$ & 0.41 & 0.48 & 0.55 \\
$\hat\rho$ & 0.33 & 0.74 & 1.19 \\
\bottomrule
\end{tabular}
\caption{Example-level standard-error ranges for the direct-prompt QQP and QNLI settings, with every standard error multiplied by $100$. The first two rows are obtained from the covariance matrix of $X_i=((B_{i1}+B_{i2})/2,B_{i1}B_{i2})$; the $\hat\rho$ row uses the delta method. With $N=8192$, both moments have standard errors below one unit on the displayed percentage scale.}
\label{tab:se-ranges-main}
\end{table}

\paragraph{Mixture calculations.}
With fixed mixture weights, a component policy is sampled independently on each call. The auxiliary mixture calculations align the recorded examples across component policies. \textcolor{black}{For component $a$, let $S_{ia}$ count correct responses among its five calls on example $i$, and let $w_a$ be its selection probability. With $\hat q_{ia}=S_{ia}/5$, the mixture uses $\hat q_{i,\mathrm{mix}}=\sum_a w_a\hat q_{ia}$ and averages $P_n(\hat q_{i,\mathrm{mix}})$ for vote curves.} These are finite-repeat plug-in summaries, not observations from executed mixture policies. \textcolor{black}{For the second moment, within-component estimates use the proportion of response pairs that are both correct; cross-component estimates multiply the two success rates.}

\paragraph{Direct-prompt QQP and QNLI settings.}
\label{app:grid-protocols-main}
For each dataset, $8{,}192$ examples are sampled uniformly without replacement from the QNLI or QQP training pool. Llama 3.1-8B, Phi-4-mini, and Qwen2.5-7B are evaluated at temperatures $0.2$, $1.0$, and $2.0$, with five calls per example. The nine model--temperature combinations define the direct-prompt settings for each dataset; Table~\ref{tab:app-all-results} gives the ID mapping. Fixed prompts request a Yes/No answer. The parser reads the JSON answer, falling back to a standalone Yes/No token; all recorded calls parse successfully, so label majority and correctness majority coincide on these records.

\paragraph{Empirical vote estimator.}
For GLUE and GSM8K, let $R$ be the number of calls and $S_i$ the number correct on example $i$. The empirical majority accuracy averages over all $K$-subsets:
\[
T_{iK}=\frac{\sum_{\ell>K/2}\binom{S_i}{\ell}\binom{R-S_i}{K-\ell}}{\binom RK},
\qquad \widehat V_{(K-1)/2}=\frac1N\sum_iT_{iK},
\]
Impossible binomial coefficients are zero, and $T_{i1}=S_i/R$. The resulting empirical accuracies are point summaries of the recorded calls. Confidence intervals obtained from two calls use Wald moment projection; the three-call checks in Appendix~\ref{app:positive-checks-main} use paired scalar Wald intervals.
\paragraph{Call accounting.}
\label{app:label-accounting-main}
Two-call validation uses $2N$ model calls and at most $N$ ground-truth labels, with each label scoring both calls. \textcolor{black}{Table~\ref{tab:se-ranges-main} reports example-level standard errors for these coordinates.}

\section{Additional gain bounds and higher-budget results}
\label{app:headroom-results-main}

\subsection{Setting N04: low-temperature Qwen decision}
For N04, the first two repeats give pair counts $(n_{00},n_{01},n_{10},n_{11})=(1272,31,20,6869)$. Let $d=\PP(B_1\ne B_2)=2(\mu-\nu)$. The polynomial inequalities
\[
P_1(q)-q\le q(1-q),\qquad
P_2(q)-q\le\frac{7\sqrt7}{18}q(1-q)
\]
imply $V_1-\mu\le d/2$ and $V_2-\mu\le(7\sqrt7/18)d/2$. Indeed, the ratios to $q(1-q)$ are $2q-1$ and $(2q-1)\{1+3q(1-q)\}$; maximizing them on $[0,1]$ gives the constants above.

The linear projection of the Wald ellipse gives
\[
d_U=\hat d+\sqrt{\chi^2_{2,1-\alpha/J}\,\hat s_d^2/N},\qquad
\hat d=\frac{51}{8192},\qquad
\hat s_d^2=\frac{N}{N-1}\hat d(1-\hat d).
\]
Here $J=1$ gives a single-condition bound, while $J=18$ accounts for choosing from the fixed 18-condition family. Combining this upper limit with the polynomial inequalities bounds both vote gains on the same moment-region event. \textcolor{black}{The bounds have nominal large-sample confidence $1-\alpha$.}

\begin{table}[H]
\centering\small
\begin{tabular}{rccc}
\toprule
Calls & Point gain interval & Wald upper & 18-setting Wald upper\\
\midrule
3 & $[-0.2990,+0.3090]$ & $0.418$ & $0.460$\\
5 & $[-0.3203,+0.3203]$ & $0.430$ & $0.474$\\
\bottomrule
\end{tabular}
\caption{Setting N04: QNLI/Qwen2.5-7B at $T=0.2$. All limits are on the absolute percentage-accuracy scale $100(V_n-V_0)$. The last two columns combine the Wald upper limit for disagreement with the polynomial gain bounds, at nominal level $1-\alpha=0.95$.}
\label{tab:qwen-decision-app}
\end{table}

The full five-repeat record gives empirical gains of $+0.011$ and $-0.029$ on the percentage-accuracy scale. \textcolor{black}{These point estimates use the subset estimator; two-call gain bounds use moment projection.}
\subsection{GSM8K evaluation and budget curves}
\label{app:gsm8k-details-main}

This supplementary higher-budget evaluation uses Qwen3-14B on GSM8K \citep{Cobbe2021GSM8K,QwenTeam2025}. Table~\ref{tab:gsm8k-main} reports the empirical accuracy curve and Wald-projected two-moment bounds.

The evaluation uses $496$ problems from a uniform draw of the official GSM8K test set, after excluding the $16$ problems used during development. Qwen3-14B generates nine fresh calls per problem at temperature $0.7$ and top-$p=0.95$, with native thinking disabled. The fixed evaluator removes commas and requires exact numeric agreement with the gold answer. All recorded responses were successfully generated and parsed.

The empirical curves use the subset estimator in Appendix~\ref{app:numerics-main}. The Wald intervals use only the first two calls, whose pooled pair counts are 32 both-wrong, 23 discordant, and 441 both-correct. All budgets use the same moment ellipse. The empirical three- and nine-vote gains over the single-call accuracy estimated from all nine calls are $+0.42$ and $+0.85$, \textcolor{black}{percentage points, respectively.}

\begin{table}[H]
\centering\small
\begin{tabular}{rcc}
\toprule
Calls $K$ & \shortstack{Empirical accuracy\\ (nine-call record)} & \shortstack{Wald-projected accuracy interval\\ (first-two-call moments)}\\
\midrule
1 & 91.89 & $[88.35,94.11]$\\
3 & 92.31 & $[86.90,96.13]$\\
5 & 92.50 & $[86.11,96.32]$\\
7 & 92.69 & $[85.62,96.46]$\\
9 & 92.74 & $[85.28,96.58]$\\
\bottomrule
\end{tabular}
\caption{GSM8K strict-majority correctness accuracies and Wald-projected intervals (\%). Empirical accuracies average all subsets of the nine-call record; the moment estimate uses the first two calls, with $\hat\mu=91.23\%$. All intervals are projections of the same nominal $95\%$ Wald region.}
\label{tab:gsm8k-main}
\end{table}

Strict-majority correctness guarantees plurality success for the unique correct numeric answer, while split wrong answers can make plurality more accurate. With numeric normalization and either uniform random or earliest-answer tie-breaking, the observed plurality curves are:
\[
\begin{array}{c|ccccc}
K & 1 & 3 & 5 & 7 & 9\\ \hline
\text{uniform-random-tie plurality (\%)} & 91.89 & 92.51 & 92.83 & 93.02 & 93.25\\
\text{first-answer-tie plurality (\%)} & 91.89 & 92.55 & 92.84 & 93.03 & 93.15
\end{array}
\]
These plurality accuracies are secondary outcomes. The strict-majority lower bound also bounds plurality accuracy from below; the upper bound applies only to strict-majority accuracy.

\subsection{Matched dependence stress diagnostic}
\label{app:dependence-stress-main}
On a matched subset of $128$ problems, a cache repeats each of three fresh generations in a block of three, intentionally violating conditional independence. The first two cached calls always agree, collapsing their plug-in interval to a point, yet the observed higher-budget accuracy can differ from that point \textcolor{black}{(by up to $0.78$ percentage points)}. \textcolor{black}{\textcolor{black}{This illustrates the role of conditional independence.}}

\section{Detailed LLM geometry and results}
\label{app:llm-table}

\textcolor{black}{Table~\ref{tab:app-all-results} identifies the 18 direct-prompt settings; Tables~\ref{tab:confirmation-limits} and \ref{tab:confirmation-checks} report Q04 and N05. Q01--Q05 and N01--N05 appear in Figure~\ref{fig:mu-rho}; the remaining settings extend the temperature comparisons. Q denotes QQP and N denotes QNLI. Q04 uses longer QQP pairs, while N05 varies the QNLI wording independently on each call. \textcolor{black}{Both rules were developed and evaluated on separate pools.}}

\begin{figure}[H]
\centering
\includegraphics[width=\linewidth]{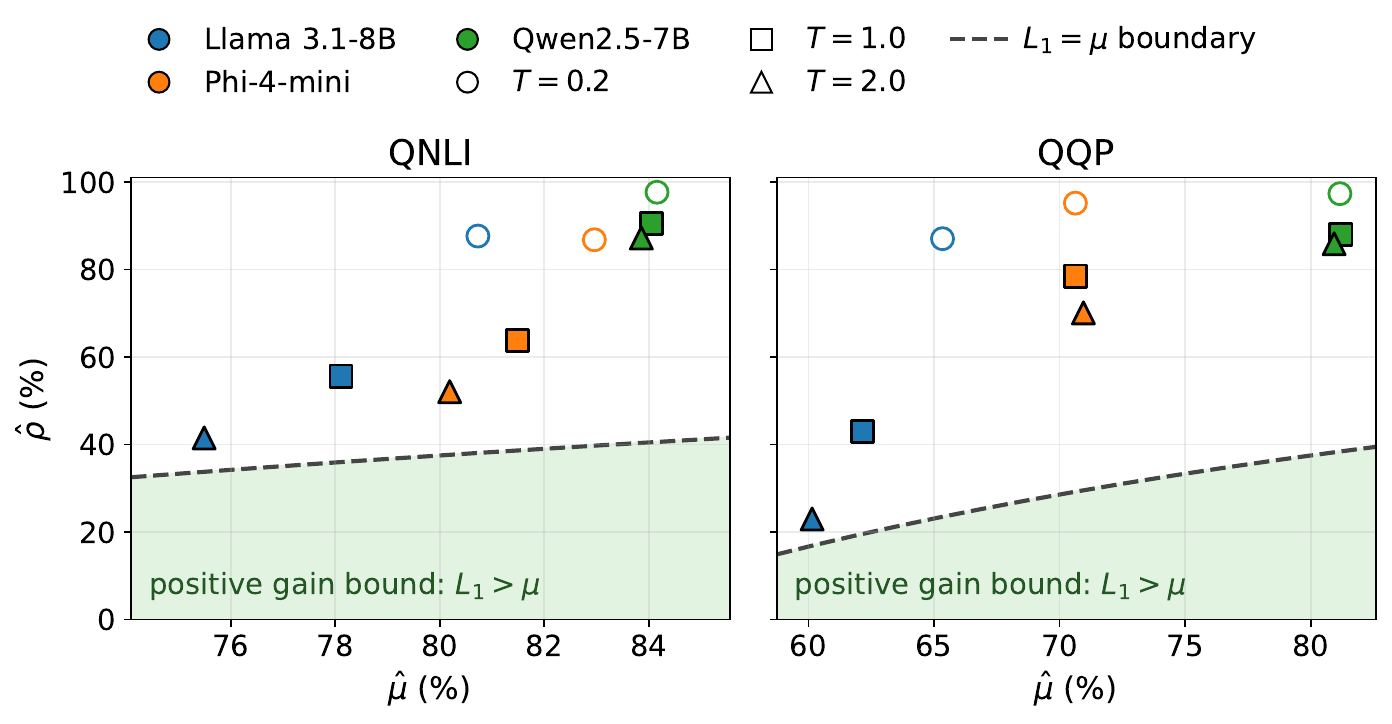}
\caption{\textcolor{black}{Direct-prompt QNLI settings (left) and QQP settings (right), by model (color) and temperature (shape), with horizontal $\hat\mu$ and vertical $\hat\rho$ in percent.} Table~\ref{tab:app-all-results} maps models and temperatures to IDs. \textcolor{black}{The dashed curve is the population boundary} $\rho=1-1/(2\mu)$, and the shaded region is $L_1>\mu$. \textcolor{black}{Figure~\ref{fig:mu-rho} shows the middle-temperature settings and the low-temperature Q05 and N04 alongside Q04 and N05, whose gains appear in its right panel;} Table~\ref{tab:app-all-results} gives the observed gains for the settings shown here.}
\label{fig:original-mu-rho}
\end{figure}

\textcolor{black}{Table~\ref{tab:app-all-results} gives settings and results;} Figure~\ref{fig:original-mu-rho} shows the direct-prompt QQP and QNLI settings \citep{Ollama2026Llama31,Ollama2026Phi4Mini,Ollama2026Qwen25}. For the 18 direct-prompt settings, the first two calls estimate the moment coordinates and confidence region; all five calls contribute to the subset-averaged empirical gains. \textcolor{black}{Every observed three- and five-vote accuracy lies inside its projected row-wise Wald $95\%$ region.} Since the calculations share calls, this is a compatibility check rather than independent coverage validation.

\begin{table}[H]
\centering
\scriptsize
\setlength{\tabcolsep}{2.7pt}
\resizebox{\textwidth}{!}{%
\begin{tabular}{llrrrrrr}
\toprule
Setting & Policy & $\hat\mu$ (\%) & $\hat\rho$ (\%) & $100(V_1-V_0)$ & $100(V_2-V_0)$ & $[L_1,U_1]_{95}$ (\%) & $[L_2,U_2]_{95}$ (\%) \\
\midrule
N06 & Llama 3.1-8B $T=0.2$ & 80.73 & 87.64 & +0.11 & +0.08 & [78.10,83.59] & [77.67,83.73] \\
N01 & Llama 3.1-8B $T=1.0$ & 78.11 & 55.63 & +1.39 & +2.03 & [74.61,85.17] & [72.92,86.78] \\
N07 & Llama 3.1-8B $T=2.0$ & 75.49 & 41.51 & +1.62 & +2.47 & [72.98,84.17] & [71.80,86.99] \\
N08 & Phi-4-mini $T=0.2$ & 82.96 & 86.79 & +0.11 & +0.08 & [80.46,85.71] & [80.01,85.85] \\
N02 & Phi-4-mini $T=1.0$ & 81.49 & 63.79 & +0.94 & +1.21 & [78.16,87.13] & [76.70,88.02] \\
N09 & Phi-4-mini $T=2.0$ & 80.19 & 52.08 & +1.34 & +1.98 & [77.29,87.25] & [75.90,88.84] \\
N04 & Qwen2.5-7B $T=0.2$ & 84.16 & 97.66 & +0.01 & -0.03 & [82.87,85.45] & [82.85,85.46] \\
N03 & Qwen2.5-7B $T=1.0$ & 84.06 & 90.53 & +0.08 & +0.17 & [81.98,86.25] & [81.73,86.32] \\
N10 & Qwen2.5-7B $T=2.0$ & 83.86 & 87.19 & +0.05 & +0.14 & [81.48,86.47] & [81.07,86.60] \\
Q06 & Llama 3.1-8B $T=0.2$ & 65.34 & 87.04 & -0.04 & -0.06 & [61.60,69.27] & [61.02,69.62] \\
Q01 & Llama 3.1-8B $T=1.0$ & 62.16 & 42.97 & +1.47 & +1.92 & [56.90,70.98] & [54.07,74.85] \\
Q07 & Llama 3.1-8B $T=2.0$ & 60.15 & 22.97 & +1.73 & +2.57 & [57.33,68.54] & [55.95,73.17] \\
Q08 & Phi-4-mini $T=0.2$ & 70.63 & 95.17 & -0.10 & -0.10 & [68.46,72.82] & [68.36,72.88] \\
Q02 & Phi-4-mini $T=1.0$ & 70.62 & 78.49 & +0.03 & -0.02 & [66.27,75.69] & [65.04,76.39] \\
Q09 & Phi-4-mini $T=2.0$ & 70.95 & 70.09 & +0.12 & +0.32 & [66.15,77.17] & [64.33,78.39] \\
Q05 & Qwen2.5-7B $T=0.2$ & 81.16 & 97.33 & +0.03 & +0.02 & [79.70,82.61] & [79.67,82.63] \\
Q03 & Qwen2.5-7B $T=1.0$ & 81.20 & 88.00 & 0.00 & -0.01 & [78.65,83.96] & [78.24,84.10] \\
Q10 & Qwen2.5-7B $T=2.0$ & 80.93 & 85.88 & +0.03 & +0.04 & [78.16,84.00] & [77.63,84.18] \\
\bottomrule
\end{tabular}}
\caption{Settings and results for the 18 direct-prompt QNLI (N) and QQP (Q) settings. Each row identifies the model, temperature, and setting ID. Each task uses $8192$ examples sampled without replacement, with five calls per example (Appendix~\ref{app:grid-protocols-main}). $V_1$ and $V_2$ denote three- and five-vote accuracy; gains are in percentage points. The moment estimates use the first two calls; empirical gains use all five calls and their single-call baseline. The last two columns are row-wise nominal $95\%$ Wald accuracy regions.}
\label{tab:app-all-results}
\end{table}

\section{Additional empirical summaries}
\label{app:additional-empirical-main}

\textcolor{black}{The 18 direct-prompt conditions also support analyses of mixtures and point predictions.}

\subsection{Randomized mixture policies}
\label{app:mixture-main}

A randomized policy would select a component independently on each call. Table~\ref{tab:mixture-gain-main} summarizes an auxiliary search for five-vote gain using the aligned plug-in estimates in Appendix~\ref{app:numerics-main}. These fitted, same-data selections require direct validation; they differ from the independently sampled prompt protocol used for the held-out QNLI gain evaluation.

\begin{table}[H]
\centering
\scriptsize
\setlength{\tabcolsep}{3pt}
\begin{tabularx}{\textwidth}{ll>{\raggedright\arraybackslash}Xrrrr}
\toprule
Dataset & Type & Mixture & $\hat\mu$ (\%) & $\hat\rho$ (\%) & $V_2$ (\%) & $100(V_2-\mu)$ \\
\midrule
QNLI & pair & 0.45$\cdot$Llama 3.1-8B (T=2.0) + 0.55$\cdot$Phi-4-mini (T=2.0) & 77.91 & 37.85 & 81.01 & +3.10 \\
QNLI & triple & 0.40$\cdot$Llama 3.1-8B (T=2.0) + 0.40$\cdot$Phi-4-mini (T=2.0) + 0.20$\cdot$Qwen2.5-7B (T=2.0) & 78.91 & 38.23 & 82.35 & +3.44 \\
QQP & pair & 0.40$\cdot$Llama 3.1-8B (T=2.0) + 0.60$\cdot$Qwen2.5-7B (T=0.2) & 72.72 & 42.43 & 77.89 & +5.17 \\
QQP & triple & 0.40$\cdot$Llama 3.1-8B (T=2.0) + 0.10$\cdot$Phi-4-mini (T=2.0) + 0.50$\cdot$Qwen2.5-7B (T=0.2) & 71.73 & 39.28 & 76.76 & +5.04 \\
\bottomrule
\end{tabularx}
\caption{Mixtures selected for largest plug-in five-vote gain on the auxiliary grid. Values describe simulated independent component selection using fitted per-example success probabilities. The gain column is a difference of percentage accuracies, not a relative improvement.}
\label{tab:mixture-gain-main}
\end{table}

\begin{figure}[H]
\centering
\includegraphics[width=0.95\textwidth]{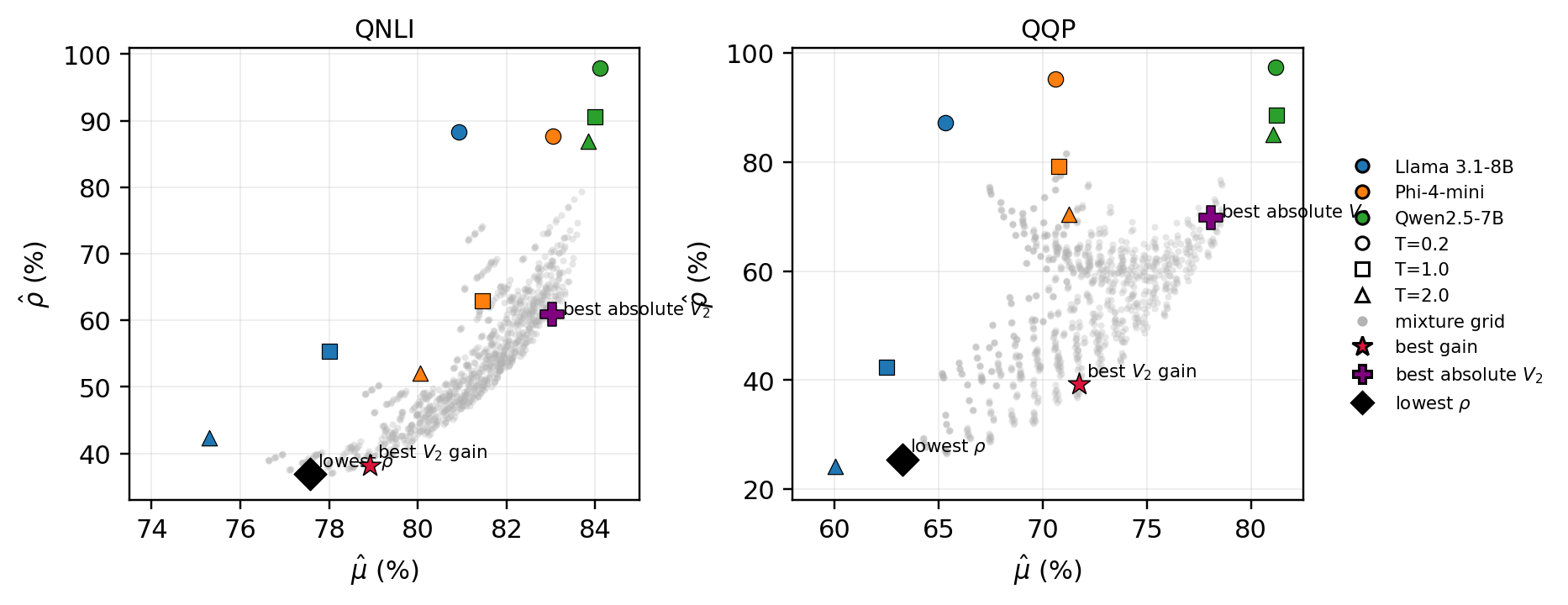}
\caption{Plug-in mixture summaries in the two-call plane, estimated from recorded base-policy responses. Gray points are mixtures and colored markers are base policies. These fitted mixtures differ from the directly evaluated QNLI prompt sampler.}
\label{fig:policy-mixture-main}
\end{figure}

\subsection{Overtaking examples}
\label{app:overtaking-main}

Table~\ref{tab:overtaking-main} gives examples where policy $A$ has lower single-call accuracy than policy $B$ but higher accuracy after voting. The differences are paired on the same example set and reported as differences of percentage accuracies, $100\{V_n(A)-V_n(B)\}$.

\begin{table}[H]
\centering
\scriptsize
\setlength{\tabcolsep}{3pt}
\begin{tabularx}{\textwidth}{l>{\raggedright\arraybackslash}X>{\raggedright\arraybackslash}Xrrr}
\toprule
Dataset & Policy $A$ & Policy $B$ & $100\Delta_0$ & $100\Delta_1$ & $100\Delta_2$ \\
\midrule
QNLI & N09: Phi-4-mini (T=2.0) & N06: Llama 3.1-8B (T=0.2) & -0.86 & +0.37 & +1.04 \\
QNLI & 0.30$\cdot$Llama 3.1-8B (T=2.0) + 0.20$\cdot$Phi-4-mini (T=1.0) + 0.50$\cdot$Qwen2.5-7B (T=1.0) & Llama 3.1-8B (T=0.2) & -0.04 & +1.95 & +2.67 \\
QQP & 0.40$\cdot$Llama 3.1-8B (T=2.0) + 0.20$\cdot$Phi-4-mini (T=2.0) + 0.40$\cdot$Qwen2.5-7B (T=0.2) & Phi-4-mini (T=1.0) & -0.05 & +3.13 & +4.56 \\
\bottomrule
\end{tabularx}
\caption{Policies with lower single-call accuracy can have higher accuracy after three or five votes. Here $\Delta_n=V_n(A)-V_n(B)$, so the displayed differences are on the percentage-accuracy scale. Mixture rows use the auxiliary plug-in calculation; base-policy comparisons use the recorded-repeat estimator.}
\label{tab:overtaking-main}
\end{table}

\subsection{Point summaries of finite-vote accuracy}
\label{app:point-completion-comparison-main}

Table~\ref{tab:point-completion-main} compares three point predictions of finite-vote accuracy against the empirical three- and five-vote estimates in the 18 base-policy runs. The interval midpoint uses the sharp bounds evaluated at the first-two-call moment estimates, separately at each budget, whereas LDGP and maximum entropy select latent laws. LDGP has the lowest mean absolute error (MAE) and root mean squared error (RMSE) at both budgets. Both parametric completions have positive average bias.

\begin{table}[H]
\centering
\small
\begin{tabular}{llrrrr}
\toprule
Target & Point summary & Bias & MAE & RMSE & Max abs. error \\
\midrule
$V_1$ & Interval midpoint & +0.34 & 0.36 & 0.61 & 1.81 \\
$V_1$ & LDGP & +0.13 & 0.21 & 0.35 & 1.06 \\
$V_1$ & maximum entropy & +1.04 & 1.04 & 1.19 & 2.14 \\
$V_2$ & Interval midpoint & +0.22 & 0.35 & 0.71 & 2.20 \\
$V_2$ & LDGP & +0.15 & 0.22 & 0.38 & 1.08 \\
$V_2$ & maximum entropy & +1.22 & 1.22 & 1.38 & 2.52 \\
\bottomrule
\end{tabular}
\caption{Point-prediction error across the direct-prompt QQP and QNLI settings. Errors are differences of percentage accuracies; positive bias denotes overprediction.}
\label{tab:point-completion-main}
\end{table}

\section{Point-completion constructions}
\label{app:completion-details-main}
The following statements give the two moment-matched laws used in Section~\ref{sec:completion-main}.

\subsection{Maximum-entropy completion}

On the real line, fixing mean and variance selects a Gaussian as the maximum-entropy law. On $[0,1]$, maximizing differential entropy relative to Lebesgue measure instead selects a quadratic-exponential density \citep{Jaynes1957}. For an interior feasible pair $(\mu,\nu)$, define
\begin{equation*}
\psi(\lambda,\kappa)=\log\int_0^1
\exp\{\lambda(q-1/2)+\kappa(q-1/2)^2\}\,\dd q,
\end{equation*}
with density
\begin{equation}
f_{\lambda,\kappa}(q)=
\exp\{\lambda(q-1/2)+\kappa(q-1/2)^2-\psi(\lambda,\kappa)\}\mathbf 1_{[0,1]}(q).
\label{eq:maxent-density-main}
\end{equation}

\begin{proposition}
\label{prop:maxent-main}
For every interior feasible $(\mu,\nu)$ there is a unique $(\lambda,\kappa)$ such that \eqref{eq:maxent-density-main} has first two moments $(\mu,\nu)$. This density uniquely maximizes differential entropy among absolutely continuous laws on $[0,1]$ with those moments. Equivalently, $(\lambda,\kappa)$ uniquely minimizes
\begin{equation*}
J_{\mu,\nu}(\lambda,\kappa)=\psi(\lambda,\kappa)-\lambda(\mu-1/2)-\kappa(\nu-\mu+1/4).
\end{equation*}
Moreover,
\[
\mathrm{sign}(\lambda)=\mathrm{sign}(\mu-1/2),
\qquad
\frac{f_{\lambda,\kappa}(q)}{f_{\lambda,\kappa}(1-q)}
=
\exp\{2\lambda(q-1/2)\}.
\]
\end{proposition}

The parameter $\lambda$ tilts the density across the decision boundary, while $\kappa$ controls concentration near the center versus the endpoints; the resulting point curve has coordinates $V_n^{\ME}=\int_0^1 P_n(q)f_{\lambda,\kappa}(q)\,\dd q$ and $V_\infty^{\ME}=\int_{1/2}^1 f_{\lambda,\kappa}(q)\,\dd q$.

\subsection{Latent-difficulty Gaussian-probit completion}

Latent-difficulty Gaussian-probit (LDGP) gives a complementary completion from the established normal-ogive/probit item-response model for binary correctness \citep{Albert1992,Lord1980}. Let $Z\sim N(0,1)$ be example difficulty and set
\begin{equation}
Q=\Phi(\eta-\gamma Z),\qquad \gamma\ge0,
\label{eq:ldgp-model-main}
\end{equation}
where $\Phi$ is the standard normal cdf. The ability parameter $\eta$ shifts the policy relative to the latent difficulty scale, and $\gamma$ controls how much example difficulty spreads the success probabilities.

\begin{samepage}
\begin{proposition}
\label{prop:ldgp-main}
For every interior feasible pair $0<\mu<1$, $\mu^2<\nu<\mu$, there is a unique LDGP law of the form \eqref{eq:ldgp-model-main} matching $(\mu,\nu)$. Writing $t=\eta/\sqrt{1+\gamma^2}$ and $r=\gamma^2/(1+\gamma^2)$, the parameters are characterized by
\[
\mu=\Phi(t),\qquad \nu=\Phi_2(t,t;r),
\]
where $\Phi_2(\cdot,\cdot;r)$ is the bivariate standard normal cdf with correlation $r$. The completed curve is
\[
V_n^{\mathrm{LDGP}}=\E\bigl[P_n\{\Phi(\eta-\gamma Z)\}\bigr],
\qquad
V_\infty^{\mathrm{LDGP}}=\Phi(\eta/\gamma)
\]
for $\gamma>0$. When $\gamma=0$, the law is constant $Q\equiv\mu$, so the endpoint is $1$, $1/2$, or $0$ according as $\mu>1/2$, $\mu=1/2$, or $\mu<1/2$.
\end{proposition}
\end{samepage}

After difficulty-scale normalization, the same two moments determine both LDGP parameters and hence every vote budget. Both completions produce coherent modeled curves inside the sharp bounds. Proofs are in Appendix~\ref{app:proofs-main}.

\begin{table}[H]
\centering
\small
\setlength{\tabcolsep}{4pt}
\begin{tabular}{rrrrrrr}
\toprule
$\rho$ & $V_1^{\ME}$ & $V_2^{\ME}$ & $V_\infty^{\ME}$ & $V_1^{\mathrm{LDGP}}$ & $V_2^{\mathrm{LDGP}}$ & $V_\infty^{\mathrm{LDGP}}$ \\
\midrule
0.05 & 77.15 & 81.25 & 97.38 & 77.23 & 81.34 & 96.33 \\
0.20 & 74.57 & 76.68 & 82.64 & 74.58 & 76.63 & 81.99 \\
0.50 & 72.64 & 73.52 & 74.95 & 71.56 & 72.08 & 73.13 \\
0.80 & 71.35 & 71.61 & 71.78 & 70.23 & 70.30 & 70.43 \\
\bottomrule
\end{tabular}
\caption{Moment-matched model curves at fixed $\mu=0.70$. Values are accuracies in percent; each completion produces its finite-vote coordinates and infinite endpoint from one selected latent law.}
\label{tab:completion-curves-main}
\end{table}

At fixed one-call accuracy in Table~\ref{tab:completion-curves-main}, lower correlation produces larger modeled voting gains in both completions. Their finite-budget predictions are relatively close, while their limiting predictions differ more: the two prescriptions allocate latent mass differently across the threshold $q=1/2$. Appendix~\ref{app:point-completion-comparison-main} compares their finite-vote predictions with the LLM results. \textcolor{black}{Completions give coherent forecasts across budgets; sharp bounds support decisions across all compatible laws.}

\section{Existing assets, licenses, and terms of use}
\label{app:asset-licenses}

Our experiments use existing benchmark datasets, model weights, and local inference software only for research evaluation. We do not redistribute third-party raw data, model weights, or software binaries in the supplementary materials.

\begin{description}
    \item[GLUE QNLI and QQP.]
    We use QNLI and QQP from GLUE \citep{Wang2019}. GLUE directs users to the original datasets and their terms. QNLI is derived from SQuAD, which is distributed under CC BY-SA 4.0. QQP is derived from the Quora Question Pairs release and is subject to the applicable Quora/Kaggle data-use terms. We use these data only for evaluation and do not redistribute raw examples. Relevant pages: \url{https://huggingface.co/datasets/nyu-mll/glue}, \url{https://rajpurkar.github.io/SQuAD-explorer/}, and \url{https://www.kaggle.com/competitions/quora-question-pairs/rules}.

    \item[Models.]
    We use \texttt{llama3.1:8b}, \texttt{phi4-mini}, and \texttt{qwen2.5:7b} through Ollama for local inference. Llama 3.1 is governed by the Llama 3.1 Community License and Acceptable Use Policy; Phi-4-mini is released under the MIT License; Qwen2.5-7B-Instruct is released under Apache-2.0. We do not redistribute any model weights. Relevant pages: \url{https://www.llama.com/llama3_1/license/}, \url{https://huggingface.co/microsoft/Phi-4-mini-instruct}, and \url{https://huggingface.co/Qwen/Qwen2.5-7B-Instruct}.

    \item[GSM8K.]
    We use the official GSM8K test data released with \citet{Cobbe2021GSM8K}. The OpenAI grade-school-math repository distributes the dataset and accompanying code under the MIT License. We do not redistribute the raw problems. Relevant pages: \url{https://github.com/openai/grade-school-math} and \url{https://github.com/openai/grade-school-math/blob/master/LICENSE}.

    \item[Qwen3-14B.]
    We use the official \texttt{Qwen/Qwen3-14B} checkpoint \citep{QwenTeam2025}, released under Apache-2.0. We do not redistribute the model weights. Relevant page: \url{https://huggingface.co/Qwen/Qwen3-14B}.

    \item[Ollama.]
    All generations are run locally through Ollama's \texttt{/api/generate} endpoint. Ollama is distributed under the MIT License, and its services are subject to its Terms of Service. Relevant pages: \url{https://github.com/ollama/ollama/blob/main/LICENSE}, \url{https://docs.ollama.com/api/generate}, and \url{https://ollama.com/terms}.
\end{description}

The supplementary materials provide manuscript and figure sources, scored responses, and reconstruction code, without redistributing third-party task text or model weights.
\clearpage

\end{document}